\documentclass[11pt,a4paper]{article}
\PassOptionsToPackage{hyphens}{url}
\usepackage[hyperref]{acl2020}

\usepackage{latexsym}
\usepackage{url}
\usepackage{adjustbox}
\usepackage{expex}
\usepackage{multirow}
\usepackage{tikz}
\usepackage{wrapfig}
\usepackage{subcaption}
\usepackage{cleveref}
\crefname{section}{\S}{\S\S}
\Crefname{section}{\S}{\S\S}
\crefname{table}{Tab.}{}
\crefname{figure}{Fig.}{}
\crefname{algorithm}{Alg.}{}
\crefname{appendix}{App.}{}
\crefname{lemma}{Lemma}{}
\Crefname{theorem}{Theorem}{}
\crefname{prop}{Proposition}{}
\crefname{cor}{Corollary}{}
\crefname{align}{}{}
\crefname{equation}{}{}

\usepackage[disable]{todonotes}
\newcommand*\iftodonotes{\if@todonotes@disabled\expandafter\@secondoftwo\else\expandafter\@firstoftwo\fi}  % defines \iftodonotes{<true>}{<false>}, thanks to https://tex.stackexchange.com/questions/126559/conditional-based-on-packageoption
\makeatother

\newcommand{\note}[4][]{\todo[author=#2,color=#3,size=\scriptsize,fancyline,caption={},#1]{#4}} % default note settings, used by macros below.
\newcommand{\ryan}[2][]{\note[#1]{ryan}{violet!40}{#2}}

\newcommand{\ran}[2][]{\note[#1]{ran}{violet!40}{#2}}
\newcommand{\kat}[2][]{\note[#1]{Kat}{blue!40}{#2}}

\newcommand{\antonis}[2][]{\note[#1]{Antonis}{magenta!40}{#2}}

\newcommand{\christo}[2][]{\note[#1]{Christo}{green!40}{#2}}

\newcommand{\mans}[2][]{\note[#1]{mans}{lightblue!40}{#2}}

\newcommand{\liz}[2][]{\note[#1]{liz}{teal!40}{#2}}

\newcommand{\eleanor}[2][]{\note[#1]{Eleanor}{teal!40}{#2}}

\newcommand{\elenaklyachko}[2][]{\note[#1]{elenaklyachko}{red!40}{#2}}

\def\checkmark{\tikz\fill[scale=0.4](0,.35) -- (.25,0) -- (1,.7) -- (.25,.15) -- cycle;} 

\newcommand{\cd}[1]{{\texttt{#1}}}

\usepackage{microtype}

\usepackage{booktabs}
\everypar{\looseness=-1}
\usepackage{xcolor}
\usepackage{colortbl}
\usepackage{float}
\definecolor{LightGray}{gray}{0.8}
\usepackage[hang,flushmargin]{footmisc}

\aclfinalcopy % Uncomment this line for the final submission

\newcommand{\md}[1]{{\texttt{#1}}}

\usepackage{tipa}
\newcommand{\jhu}{\textrm{\normalfont \textipa{Z}}}
\newcommand{\cmu}{\textrm{\normalfont \textbeltl}}
\newcommand{\york}{\textrm{\normalfont \textipa{7}}}
\newcommand{\ucam}{ \textrm{\normalfont \textipa{Q}}}
\newcommand{\ethz}{\textrm{\normalfont \textipa{D}}}
\newcommand{\fb}{\textrm{\normalfont \textipa{F}}}
\newcommand{\umelb}{\textrm{\normalfont \textipa{@}}}
\newcommand{\hse}{\textrm{\normalfont \textipa{E}}}
\newcommand{\cub}{\textrm{\normalfont \textipa{X}}}
\newcommand{\msu}{\textrm{\normalfont \textipa{M}}}
\newcommand{\krc}{\textrm{\normalfont \textipa{K}}}
\newcommand{\ubc}{\textrm{\normalfont \texthtb}}
\newcommand{\indu}{\textrm{\normalfont \textipa{I}}}
\newcommand{\ulou}{\textrm{\normalfont \textipa{L}}}
\newcommand{\google}{\textrm{\normalfont \textipa{5}}}

\title{SIGMORPHON 2020 Shared Task 0: Typologically Diverse Morphological Inflection}
\vspace{-.5cm}
\author{\textbf{Ekaterina Vylomova}$^\umelb$~\;~\textbf{Jennifer White}$^\ucam$~\;~\textbf{Elizabeth Salesky}$^\jhu$~\;~\textbf{Sabrina J. Mielke}$^\jhu$~\;~\textbf{Shijie Wu}$^\jhu$ \\
\textbf{Edoardo Ponti}$^\ucam$~\;~\textbf{Rowan Hall Maudslay}$^\ucam$~\;~\textbf{Ran Zmigrod}$^\ucam$~\;~\textbf{Josef Valvoda}$^\ucam$~\;~\textbf{Svetlana Toldova}$^\hse$ \\ \textbf{Francis Tyers}$^{\indu,\hse}$~\;~\textbf{Elena Klyachko}$^\hse$~\;~\textbf{Ilya Yegorov}$^\msu$~\;~\textbf{Natalia Krizhanovsky}$^\krc$~\;~\textbf{Paula Czarnowska}$^\ucam$ \\ \textbf{Irene Nikkarinen}$^\ucam$~\;~
\textbf{Andrew Krizhanovsky}$^\krc$~\;~\textbf{Tiago Pimentel}$^\ucam$~\;~\textbf{Lucas Torroba Hennigen}$^\ucam$ \\ \textbf{Christo Kirov}$^\google$~\;~\textbf{Garrett Nicolai}$^\ubc$~\;~
\textbf{Adina Williams}$^\fb$~\;~\textbf{Antonios Anastasopoulos}$^\cmu$ \\ \textbf{Hilaria Cruz}$^\ulou$~\;~\textbf{Eleanor Chodroff}$^\york$~\;~\textbf{Ryan Cotterell}$^\ucam$,$^\ethz$~\;~\textbf{Miikka Silfverberg}$^\ubc$~\;~\textbf{Mans Hulden}$^\cub$ \\
$^\umelb$University of Melbourne $^\ucam$University of Cambridge $^\jhu$Johns Hopkins University \\
$^\hse$Higher School of Economics $^\msu$Moscow State University $^\krc$Karelian Research Centre \\
$^\google$Google AI $^\ubc$University of British Columbia $^\fb$Facebook AI Research \\
$^\cmu$Carnegie Mellon University
$^\indu$Indiana University $^\ulou$University of Louisville \\ $^\york$University of York $^\ethz$ETH Z{\"u}rich $^\cub$University of Colorado Boulder \\
\texttt{ekaterina.vylomova@unimelb.edu.au} \hspace{.35cm}\texttt{ryan.cotterell@inf.ethz.ch}
}

\date{}

\begin{document}
\maketitle

\begin{abstract}
\vspace{-0.1cm}
A broad goal in natural language processing (NLP) is to develop a system that has the capacity to process \textit{any} natural language. Most systems, however, are developed using data from just one language such as English. %, and typically one that is high resource, such as English. 
%Any one language reflects just one linguistic system from a variable cross-linguistic set. 
The SIGMORPHON 2020 shared task on morphological reinflection aims to investigate systems' ability to generalize across typologically distinct languages, many of which are low resource.
%If a system is developed using a sample of Indo-European languages, how well would that system work for typologically distinct languages that may differ substantially in their morphological structure? 
%Does the proposed architecture have the flexibility to overcome these morphological differences? 
%In this task, we assembled morphological paradigms for xxx languages from xxx language families using existing data from UniMorph, as well as newly contributed annotations. 
%As in previous iterations of the task, the goal was to derive the correct inflected form from a lemma and a set of morphological features. 
%In the Development Phase, participants were first presented with xxx languages from just 5 families to develop their systems.
%In the Generalization Phase, surprise languages from 3 "seen" families and 10 new families were presented for participants to fine-tune their models.
%In the Evaluation Phase, test splits for all 90 languages were released, which included data from an additional xxx language families. 
Systems were developed using data from 45 languages and just 5 language families, fine-tuned with data from an additional 45 languages and 10 language families (13 in total), and evaluated on all 90 languages.
A total of 22 systems (19 neural) from 10 teams were submitted to the task.
All four winning systems were neural (two monolingual transformers and two massively multilingual RNN-based models with gated attention). Most teams demonstrate utility of data hallucination and augmentation, ensembles, and multilingual training for low-resource languages. Non-neural learners and manually designed grammars showed competitive and even superior performance on some languages (such as Ingrian, Tajik, Tagalog, Zarma, Lingala), especially with very limited data.
Some language families (Afro-Asiatic, Niger-Congo, Turkic) were relatively easy for most systems and achieved over 90\%  mean accuracy while others %(Austronesian, Tungusic, Uto-Aztecan, Oto-Manguean, Southern Daly) 
were more challenging.
%Any quick summary of the results / errors?
\end{abstract}
\section{Introduction}
Human language is marked by considerable diversity around the world.
%\ryan{Words cannot express how much I dislike this sentence.}\kat{ I tried to express the idea that language variations/change are to a certain extent random but there are always constraints (e.g., functional adaptive) that lead to cross-linguistic regularities/universal. Or should we mention innate UG as well?}
Though the world's languages share many basic attributes (e.g., \citealp{swadesh1950salish} and more recently, \citealp{list2016concepticon}), grammatical features, and even abstract implications (proposed in \citealp{greenberg1963universals}), each language nevertheless has a unique evolutionary trajectory that is affected by geographic, social, cultural, and other factors. As a result, the surface form of languages varies substantially.
%\eleanor{taking a stab at this, but feel free to delete -- I'm stealing some of the content from the "Meet Our Families" intro}
The morphology of languages can differ in many ways: Some exhibit rich grammatical case systems (e.g., 12 in Erzya and 24 in Veps) and mark possessiveness, others might have complex verbal morphology (e.g., Oto-Manguean languages;  \citealp{palancar2016tone}) or even ``decline'' nouns for tense (e.g., Tupi--Guarani languages). Linguistic typology is the discipline that studies these variations by means of a systematic comparison of languages \cite{croft2002typology,comrie1989language}.
Typologists have defined several dimensions of morphological variation to classify and quantify the degree of cross-linguistic variation. 
This comparison can be challenging as the categories are based on studies of known languages and are progressively refined with documentation of new languages~\cite{haspelmath2007pre}.
Nevertheless, to understand the potential range of morphological variation, we take a closer look at three dimensions here: fusion, inflectional synthesis, and position of case affixes~\cite{wals}. 

Fusion, our first dimension of variation, refers to the degree to which morphemes bind to one another in a phonological word \cite{wals-20}. Languages range from strictly isolating (i.e., each morpheme is its own phonological word) to concatenative (i.e., morphemes bind together within a phonological word); non-linearities such as ablaut or tonal morphology can also be present. 
From a geographic perspective, isolating languages are found in the Sahel Belt in West Africa, Southeast Asia and the Pacific.
Ablaut--concatenative morphology and tonal morphology can be found in African languages.\ryan{This sentence is bizarre. Ablaut is also in Germaic languages, right? Also ablaut is by definition non-concatenative.}
Tonal--concatenative morphology can be found in Mesoamerican languages (e.g., Oto-Manguean).
Concatenative morphology is the most common system and can be found around the world. 
Inflectional synthesis, the second dimension considered, refers to whether grammatical categories like tense, voice or agreement are expressed as affixes (synthetic) or individual words (analytic) \cite{wals-22}. 
Analytic expressions are common in Eurasia (except the Pacific Rim, and the Himalaya and Caucasus mountain ranges), whereas synthetic expressions are used to a high degree in the Americas.
Finally, affixes can variably surface as prefixes, suffixes, infixes, or circumfixes \cite{wals-51}. Most Eurasian and Australian languages strongly favor suffixation, and the same holds true, but to a lesser extent, for South American and New Guinean languages \cite{wals-51}. 
In Mesoamerican languages and African languages spoken below the Sahara, prefixation is dominant instead.

These are just three dimensions of variation in morphology, and the cross-linguistic variation is already considerable. 
Such cross-lingual variation makes the development of natural language processing (NLP) applications challenging. 
As \citet{bender2009linguistically,bender2016linguistic} notes, many current architectures and training and tuning algorithms still present language-specific biases. 
The most commonly used language for developing NLP applications is English. 
Along the above dimensions, English is productively concatenative, a mixture of analytic and synthetic,\ryan{What about fusion?} and largely suffixing in its inflectional morphology.
With respect to languages that exhibit inflectional morphology, English is relatively impoverished.\footnote{Note that many languages exhibit no inflectional morphology e.g., Mandarin Chinese, Yoruba, etc.: \citet{wals-21}.}
Importantly, English is just one morphological system among many. 
A larger goal of natural language processing is that the system work for \emph{any} presented language.
If an NLP system is trained on just one language, it could be missing important flexibility in its ability to account for cross-linguistic morphological variation.

In this year's iteration of the SIGMORPHON shared task on morphological reinflection, we specifically focus on typological diversity and aim to investigate systems' ability to generalize across typologically distinct languages many of which are low-resource. For example, if a neural network architecture works well for a sample of Indo-European languages, should the same architecture also work well for Tupi--Guarani languages (where nouns are ``declined'' for tense) or Austronesian languages (where verbal morphology is frequently prefixing)?

%\Eleanor{@Kat: I moved the original intro to notes.tex. Feel free to move back if you'd prefer!}
%\Kat{@Eleanor, thank you! The reworked intro reads much better now!}

\section{Task Description}
%A %visual stimulus for the task is found in \cref{fig:wug}. 
%\begin{figure}
%\centering
%    \includegraphics[width=0.2\textwidth]{figures/WugTestC.jpg}
%    \caption{Demonstration of task in \citealp{berko1958child}. }
%    \label{fig:wug}
%    \end{figure}
The 2020 iteration of our task is similar to CoNLL-SIGMORPHON 2017 \cite{cotterell-etal-2017-conll}  and 2018 \cite{cotterell2018conll} in that participants are required to design a model that learns to generate inflected forms from a lemma and a set of morphosyntactic features that derive the desired target form. For each language we provide a separate training, development, and test set. More historically, all of these tasks resemble the classic ``wug''-test that \newcite{berko1958child} developed to test child and human knowledge of English nominal morphology.\looseness=-1

Unlike the task from earlier years, this year's task proceeds in three phases: a Development Phase, a Generalization Phase, and an Evaluation Phase, in which each phase introduces previously unseen data.
The task starts with the \textbf{Development Phase}, which was an elongated period of time (about two months), during which participants \emph{develop} a model of morphological inflection. In this phase, we provide training and development splits for 45 languages representing the Austronesian, Niger-Congo, Oto-Manguean, Uralic and Indo-European language families. Table~\ref{tab:lang-descr} provides details on the languages.
The \textbf{Generalization Phase} is a short period of time (it started about a week before the Evaluation Phase) during which participants fine-tune their models on new data. At the start of the phase, we provide training and development splits for 45 new languages where approximately half are genetically related (belong to the same family) and half are genetically unrelated (are isolates or belong to a different family) to the languages presented in the Development Phase. More specifically, we introduce (surprise) languages from Afro-Asiatic, Algic, Dravidian, Indo-European, Niger-Congo, Sino-Tibetan, Siouan, Songhay, Southern Daly, Tungusic, Turkic, Uralic, and Uto-Aztecan families. See Table~\ref{tab:lang-descr2} for more details.

Finally, test splits for all 90 languages are released in the \textbf{Evaluation Phase}. During this phase, the models are evaluated on held-out forms. Importantly, the languages from both previous phases are evaluated simultaneously. This way, we evaluate the extent to which models (especially those with shared parameters) overfit to the development data: a model based on the morphological patterning of the Indo-European languages may end up with a bias towards suffixing and will struggle to learn prefixing or infixation. 

\begin{figure*}
    \centering
    \includegraphics[width=\textwidth]{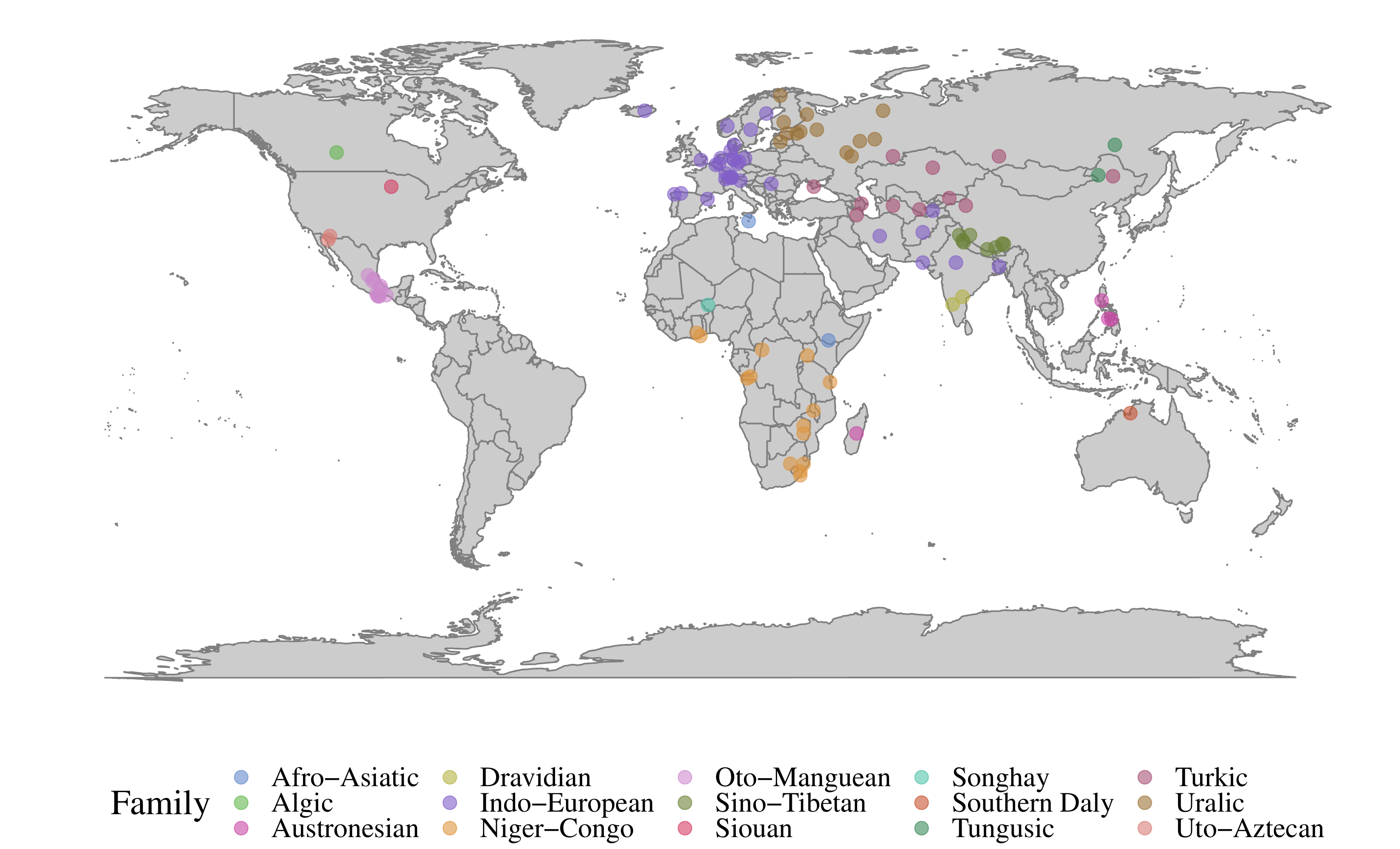}
    \caption{Languages in our sample colored by family.}
    \label{fig:map}
\end{figure*}

\section{Meet our Languages}
In the 2020 shared task we cover 15 language families: Afro-Asiatic, Algic, Austronesian, Dravidian, Indo-European, Niger-Congo, Oto-Manguean, Sino-Tibetan, Siouan, Songhay, Southern Daly, Tungusic, Turkic, Uralic, and Uto-Aztecan.\footnote{The data splits are available at \url{https://github.com/sigmorphon2020/task0-data/}} 
Five language families were used for the Development phase while ten were held out for the Generalization phase. \cref{tab:lang-descr} and \cref{tab:lang-descr2} provide information on the languages, their families, and sources of data.
In the following section, we provide an overview of each language family's morphological system.

\subsection{Afro-Asiatic}
%%Irene(?)
The Afro-Asiatic language family, consisting of six branches and over 300 languages, is among the largest language families in the world. It is mainly spoken in Northern, Western and Central Africa as well as West Asia and spans large modern languages such as Arabic, in addition to ancient languages like Biblical Hebrew. Similarly, some of its languages have a long tradition of written form, while others have yet to incorporate a writing system. 
The six branches differ most notably in typology and syntax, with the Chadic language being the main source of differences, which has sparked discussion of the division of the family \cite{Frajzyngier2018}. For example, in the Egyptian and Semitic branches, the root of a verb may not contain vowels, while this is allowed in Chadic. Although only four of the six branches, excluding Chadic and Omotic, use a prefix and suffix in conjugation when adding a subject to a verb, it is considered an important characteristic of the family. In addition, some of the families in the phylum use tone to encode tense, modality and number among others. However, all branches use objective and passive suffixes. Markers of tense are generally simple, whereas aspect is typically distinguished with more elaborate systems.

\subsection{Algic}
%\Eleanor{Cree. Concatenative and suffixing. Distinction between impersonal and non-impersonal verbs. Four apparent declension classes among non-impersonal verbs.}
The Algic family embraces\ryan{Bold verb choice} languages native to North America---more specifically the United States and Canada---and contain three branches. Of these, our sample contains Cree, the language from the largest genus, Algonquian, from which most languages are now extinct. 
The Algonquian genus is characterized by its concatenative morphology.
Cree morphology is also concatenative and suffixing. It distinguishes between impersonal and non-impersonal verbs and presents four apparent declension classes among non-impersonal verbs.

%%Eleanor
\subsection{Austronesian}
The Austronesian family of languages is largely comprised of languages from the Greater Central Philippine and Oceanic regions.
They are characterized\ran{unclear if we are using -ise or ize?} by limited morphology, mostly prefixing in nature.
Additionally, tense--aspect affixes are predominantly seen as prefixes, though some suffixes are used.
In the general case, verbs do not mark number, person, or gender.
In M\=aori, verbs may be suffixed with a marker indicating the passive voice.
% Jennifer
This marker takes the form of one of twelve endings.
These endings are difficult to predict as the language has undergone a loss of word-final consonants and there is no clear link between a stem and the passive suffix that it employs \citep{harlow2007maori}.

%Moved to error analysis
%\antonis{This paragraph belongs in the Results/Analysis section?}

%%Jennifer
\subsection{Dravidian}
The family of Dravidian languages comprises several languages which are primarily spoken across Southern India and Northern Sri Lanka, with over 200 million speakers. The shared task includes Kannada and Telugu.
Dravidian languages primarily use the SOV word order.
They are agglutinative, and primarily use suffixes.
A Dravidian verb indicates voice, number, tense, aspect, mood and person, through the affixation of multiple suffixes.
Nouns indicate number, gender and case.
%%Kat: Temporary put the table here 
%%This should probably go into Appendix + reformatted to fit the page + we need to add train/dev/test split size 
%% Sources should be moved to references
%% CODES Are ISO 639-3; Family/Genus -- based on WALS
\begin{table*}[ht]
\begin{adjustbox}{width=1\textwidth}
\small
\begin{tabular}{l|l|l|l|l}
%%DEVELOPMENT

\multicolumn{5}{l}{\textbf{Development}}\\
 Family & Genus & ISO 639-3 & Language & Source of Data \\
\toprule
Austronesian   &Barito   &mlg (plt)   &Malagasy    &\citet{malagasy} \\
    &Greater Central Philippine   &ceb   &Cebuano  & \citet{cebuano}            \\
    &Greater Central Philippine  &hil  & Hiligaynon  & \citet{hiligaynon} \\
    &Greater Central Philippine  &tgl   &Tagalog     & \citet{tagalog}     \\
    &Oceanic  &mao (mri)  & M\=aori      & \citet{maori}           \\
    \midrule
Indo-European  & Germanic   &ang   &Old English & \href{https://unimorph.github.io/}{UniMorph}\\
    &Germanic   &dan   &Danish & \href{https://unimorph.github.io/}{UniMorph} \\
    &Germanic    &deu   &German & \href{https://unimorph.github.io/}{UniMorph}\\
    &Germanic    &eng   &English& \href{https://unimorph.github.io/}{UniMorph}\\
    &Germanic    &frr   &North Frisian    & \href{https://unimorph.github.io/}{UniMorph}\\
    &Germanic    &gmh   & Middle High German      & \href{https://unimorph.github.io/}{UniMorph}\\
    &Germanic  &isl   & Icelandic & \href{https://unimorph.github.io/}{UniMorph}\\
    &Germanic    &nld   &Dutch& \href{https://unimorph.github.io/}{UniMorph} \\
    &Germanic     &nob   &Norwegian Bokmål      & \href{https://unimorph.github.io/}{UniMorph}\\
    &Germanic     &swe   &Swedish      & \href{https://unimorph.github.io/}{UniMorph}\\
    \midrule
Niger-Congo &Bantoid   &kon (kng)   &Kongo& \citet{kongo}  \\
    &Bantoid    &lin   &Lingala & \citet{lingala}\\
    &Bantoid   &lug   &Luganda     &\citet{luganda}\\
    &Bantoid    &nya   &Chewa  & \citet{chewa}\\
    &Bantoid &sot   &Sotho & \citet{sotho}         \\
    &Bantoid    &swa (swh)   & Swahili &\citet{swahili}    \\
    &Bantoid    &zul   &Zulu      & \citet{zulu}       \\
    &Kwa   &aka   &Akan & \citet{akan} \\
    &Kwa   &gaa   &Gã       & \citet{ga} \\
    \midrule
Oto-Manguean    &Amuzgoan  &azg   &San Pedro Amuzgos Amuzgo      & \citet{surrey}        \\
    &Chichimec  &pei   &Chichimeca-Jonaz      & \citet{surrey}     \\
    &Chinantecan   &cpa  &Tlatepuzco Chinantec       & \citet{surrey}         \\
    &Mixtecan   &xty   &Yoloxóchitl Mixtec      & \citet{surrey}            \\
    &Otomian   &ote   &Mezquital Otomi  & \citet{surrey}            \\
    &Otomian  &otm   &Sierra Otomi  & \citet{surrey}    \\
    &Zapotecan    &cly   &Eastern Chatino of San Juan Quiahije &    \citet{cruz-anastasopoulos-stump:2020:LREC}    \\
    &Zapotecan &ctp   & Eastern Chatino of Yaitepec   & \citet{surrey}     \\
    &Zapotecan  &czn   &Zenzontepec Chatino  & \citet{surrey}          \\
    &Zapotecan    &zpv   &Chichicapan Zapotec      & \citet{surrey}     \\
    \midrule
Uralic    &Finnic    &est   &Estonian   & \href{https://unimorph.github.io/}{UniMorph}          \\
    &Finnic    &fin   &Finnish     & \href{https://unimorph.github.io/}{UniMorph}           \\
    &Finnic    &izh   &Ingrian      & \href{https://unimorph.github.io/}{UniMorph}            \\
    &Finnic   &krl   &Karelian     &\citet{zaytseva2017open}  \\
    &Finnic  &liv   &Livonian       & \href{https://unimorph.github.io/}{UniMorph}           \\
    &Finnic   &vep   &Veps  &\citet{zaytseva2017open}  \\
    &Finnic   &vot   &Votic  & \href{https://unimorph.github.io/}{UniMorph}\\
    &Mari   &mhr   &Meadow Mari    &\citet{arkhangeskiy2012creation}\\
    &Mordvin    &mdf   &Moksha     &\citet{arkhangeskiy2012creation}\\
    &Mordvin    &myv   &Erzya    &\citet{arkhangeskiy2012creation} \\
    &Saami   &sme   &Northern Sami       & \href{https://unimorph.github.io/}{UniMorph}   \\      
\end{tabular}
\end{adjustbox}
\caption{Development languages used in the shared task.}
\label{tab:lang-descr}
\end{table*}

\subsection{Indo-European}
Languages in the Indo-European family are native to most of Europe and a large part of Asia---with our sample including languages from the genera: Germanic, Indic, Iranian, and Romance. This is (arguably) the most well studied language family, containing a few of the highest-resource languages in the world.

\paragraph{Romance}
The Romance genus comprises of a set of fusional languages evolved from Latin.
They traditionally originated in Southern and Southeastern Europe, though they are presently spoken in other continents such Africa and the Americas.
Romance languages mark tense, person, number and mood in verbs,
and gender and number in nouns.
Inflection is primarily achieved through suffixes, with some verbal person syncretism and suppletion for high-frequency verbs.
There is some morphological variation within the genus, such as French, which exhibits comparatively less inflection, and Romanian has comparatively more---it still marks case.

\paragraph{Germanic}
The Germanic genus comprises several languages which originated in Northern and Northwestern Europe, and today are spoken in many parts of the world.
Verbs in Germanic languages mark tense and mood, in many languages person and number are also marked, predominantly through suffixation.
Some Germanic languages exhibit widespread Indo-European ablaut.
The gendering of nouns differs between Germanic languages: German nouns can be masculine, feminine or neuter, while English nouns are not marked for gender.  In Danish and Swedish, historically masculine and feminine nouns have merged to form one common gender, so nouns are either common or neuter.
Marking of case also differs between the languages: German nouns have one of four cases and this case is marked in articles and adjectives as well as nouns and pronouns, while English does not mark noun case (although Old English, which also appears in our language sample, does).

\paragraph{Indo-Iranian}
The Indo-Iranian genus contains languages spoken in Iran and across the Indian subcontinent.
Over 1.5 billion people worldwide speak an Indo-Iranian language.
Within the Indo-European family, Indo-Iranian languages belong to the Satem group of languages.
Verbs in Indo-Iranian languages indicate tense, aspect, mood, number and person.
In languages such as Hindi verbs can also express levels of formality.
Noun gender is present in some Indo-Iranian languages, such as Hindi, but absent in languages such as Persian.
Nouns generally are marked for case.

\begin{table*}[!ht]
\begin{adjustbox}{width=1\textwidth}
\small
\begin{tabular}{l|l|l|l|l}
%%SURPRISE
\multicolumn{5}{l}{\textbf{Generalization (Surprise)}}\\
 Family & Genus & ISO 639-3 & Language & Source of Data \\
\toprule
 Afro-Asiatic & Semitic & mlt & Maltese & \href{https://unimorph.github.io/}{UniMorph} \\
& Lowland East Cushitic & orm & Oromo& \citet{oromo}  \\
 & Semitic & syc & Syriac & \href{https://unimorph.github.io/}{UniMorph} \\
\midrule
 Algic & Algonquian& cre & Plains Cree & \citet{cree} \\
 \midrule
 Tungusic & Tungusic & evn & Evenki & \citet{klyachko2020lowresourceeval}  \\
 \midrule
  Turkic & Turkic & aze (azb) & Azerbaijani & \href{https://unimorph.github.io/}{UniMorph}\\
  & Turkic & bak & Bashkir&  \href{https://unimorph.github.io/}{UniMorph}  \\
  & Turkic & crh & Crimean Tatar &  \href{https://unimorph.github.io/}{UniMorph} \\
  & Turkic & kaz & Kazakh & \citet{kazakh,kazakh2}  \\
  & Turkic  & kir & Kyrgyz & \citet{kyrgyz}   \\
  & Turkic & kjh & Khakas & \href{https://unimorph.github.io/}{UniMorph}\\
  & Turkic & tuk & Turkmen & \citet{turkmen,turkmen2}  \\
    & Turkic & uig & Uyghur & \citet{uyghur}  \\
    & Turkic & uzb & Uzbek & \citet{uzbek,uzbek2}   \\
    \midrule
 Dravidian & Southern Dravidian & kan & Kannada & \href{https://unimorph.github.io/}{UniMorph} \\
        &South-Central Dravidian& tel & Telugu & \href{https://unimorph.github.io/}{UniMorph}     \\
        \midrule
 Indo-European  & Indic & ben & Bengali & \href{https://unimorph.github.io/}{UniMorph}\\
    & Indic & hin & Hindi & \href{https://unimorph.github.io/}{UniMorph}  \\
    & Indic & san & Sanskrit & \href{https://unimorph.github.io/}{UniMorph}\\
    & Indic & urd & Urdu& \href{https://unimorph.github.io/}{UniMorph}\\
    & Iranian & fas (pes) & Persian & \href{https://unimorph.github.io/}{UniMorph} \\
    & Iranian & pus (pst) & Pashto & \href{https://unimorph.github.io/}{UniMorph} \\
    & Iranian & tgk & Tajik& \href{https://unimorph.github.io/}{UniMorph} \\
    & Romance & ast & Asturian & \href{https://unimorph.github.io/}{UniMorph} \\
    & Romance & cat & Catalan & \href{https://unimorph.github.io/}{UniMorph} \\
    & Romance & frm & Middle French & \href{https://unimorph.github.io/}{UniMorph}\\
    & Romance & fur & Friulian& \href{https://unimorph.github.io/}{UniMorph}\\
    & Romance & glg & Galician & \href{https://unimorph.github.io/}{UniMorph} \\
    & Romance & lld & Ladin & \href{https://unimorph.github.io/}{UniMorph} \\
    & Romance & vec & Venetian & \href{https://unimorph.github.io/}{UniMorph} \\
    & Romance & xno & Anglo-Norman & \href{https://unimorph.github.io/}{UniMorph}  \\
   & West Germanic & gml & Middle Low German & \href{https://unimorph.github.io/}{UniMorph} \\
    & West Germanic & gsw &Swiss German & \citet{ch}\\
    & North Germanic & nno & Norwegian Nynorsk & \href{https://unimorph.github.io/}{UniMorph} \\
 \midrule
 Niger-Congo & Bantoid & sna & Shona & \citet{shona,shona2} \\
 \midrule
Sino-Tibetan & Bodic & bod & Tibetan & \citet{di2019modelling} \\
\midrule
 Siouan & Core Siouan & dak  & Dakota & \citet{dakota}  \\
 \midrule
 Songhay & Songhay & dje &Zarma& \citet{zarma}\\
 \midrule
 Southern Daly & Murrinh-Patha & mwf & Murrinh-Patha & \citet{mansfield2019murrinhpatha}\\
 \midrule
 Uralic & Permic & kpv & Komi-Zyrian & \citet{arkhangeskiy2012creation}  \\
    & Finnic & lud & Ludic& \citet{zaytseva2017open}  \\
    & Finnic & olo & Livvi & \citet{zaytseva2017open} \\
    & Permic & udm & Udmurt & \citet{arkhangeskiy2012creation}  \\
    & Finnic & vro & V\~{o}ro & \citet{iva2007voru}\\
    \midrule
 Uto-Aztecan & Tepiman & ood & O'odham  &   \citet{oodham}
\end{tabular}
\end{adjustbox}
\caption{Surprise languages used in the shared task.}
\label{tab:lang-descr2}
\end{table*}

\subsection{Niger--Congo}
Our language sample includes two genera from the Niger--Congo family, namely Bantoid and Kwa languages. These have mostly exclusively concatenative fusion, and single exponence in verbal tense--aspect--mood. The inflectional synthesis of verbs is moderately high, e.g.\ with 4-5 classes per word in Swahili and Zulu. The locus of marking is inconsistent (it falls on both heads and dependents), and most languages are are predominantly prefixing. Full and partial reduplication is attested in most languages. Verbal person--number markers tend to be syncretic.

As for nominal classes, Bantoid languages are characterized by a large amount of grammatical genders (often more than 5) assigned based on both semantic and formal rules, whereas some Akan languages (like Ewe) lack a gender system. Plural tends to be always expressed by affixes or other morphological means. Case marking is generally absent or minimal. As for verbal classes, aspect is grammaticalized in Akhan (Kwa) and Zulu (Bantoid), but not in Luganda and Swahili (Bantoid). Both past and future tenses are inflectional in Bantoid languages. 2-3 degrees of remoteness can be distinguished in Zulu and Luganda, but not in Swahili. On the other hand, Akan (Kwa) has no opposition between past and non-past. There are no grammatical evidentials.

%% Hilaria
\subsection{Oto-Manguean}
The Oto-Manguean languages are a diverse family of tonal languages spoken in central and southern Mexico. Even though all of these languages are tonal, the tonal system within each language varies widely. Some have an inventory of two tones (e.g., Chichimec and Pame) others have ten tones (e.g., the Eastern Chatino languages of the Zapotecan branch, \citet{palancar2016tone}). 

Oto-Manguean languages are also rich in tonal morphology. The inflectional system marks person--number and aspect in verbs and person--number in adjectives and noun possessions, relying heavily on tonal contrasts. Other interesting aspects of Oto-Manguean languages include the fact that pronominal inflections use a system of enclitics, and first and second person plural has a distinction between exclusive and inclusive \cite{campbell2016tone}.
Tone marking schemes in the writing systems also vary greatly. Some writing systems do not represent tone, others use diacritics, and others represent tones with numbers. In languages that use numbers, single digits represent level tones and double digits represent contour tones. For example, in San Juan Quiahije of Eastern Chatino number 1 represents high tone, number 4 represents low tone, and numbers 14 represent a descending tone contour and numbers 42 represent an ascending tone contour \citet{cruz2014linguistic}.

\subsection{Sino-Tibetan} The Sino-Tibetan family is represented by the Tibetan language. Tibetan uses an abugida script and contains complex syllabic components in which vowel marks can be added above and below the base consonant. Tibetan verbs are inflected for tense and mood. Previous studies on Tibetan morphology \cite{di2019modelling} indicate that the majority of mispredictions produced by neural models are due to allomorphy. This is followed by generation of nonce words (impossible combinations of vowel and consonant components). \\

\subsection{Siouan}
%\Eleanor{}
The Siouan languages are located in North America, predominantly along the Mississippi and Missouri Rivers and in the Ohio Valley. The family is represented in our task by Dakota, a critically endangered language spoken in North and South Dakota, Minnesota, and Saskatchewan. The Dakota language is largely agglutinating in its derivational morphology and fusional in its inflectional morphology with a mixed affixation system \cite{RankinEtAl2003}. The present task includes verbs, which are marked for first and second person, number, and duality. All three affixation types are found: person was generally marked by an infix, but could also appear as a prefix, and plurality was marked by a suffix. Morphophonological processes of fortition and vowel lowering are also present.
%Choice of morpheme subject to semantics of the verb (looked like ergativity or subject was agent vs experiencer). 
%this maybe should have been marked in unimorph, but it wasn't mentioned in the book, and I couldn't fully figure out the process.

%%Eleanor
\subsection{Songhay} The Songhay family consists of around eleven or twelve languages spoken in Mali, Niger, Benin, Burkina Faso and Nigeria. In the shared task we use Zarma, the most widely spoken Songhay language.\antonis{Added this sentence, please confirm that this is the correct name of the language.}\kat{Yes} Most of the Songhay languages are predominantly SOV with medium-sized consonant inventories (with implosives), five phonemic vowels, vowel length distinctions, and word level tones, which also are used to distinguish nouns, verbs, and adjectives~\citep{heath2014grammar}.

\subsection{Southern Daly} The Southern Daly is a small language family of the Northern Territory in Australia that consists of two distantly related languages. In the current task we only have one of the languages, Murrinh-patha (which was initially thought to be a language isolate). Murrinh-patha is classified as polysynthetic with highly complex verbal morphology. Verbal roots are surrounded by prefixes and suffixes that indicate tense, mood, object, subject. As \citet{mansfield2019murrinhpatha} notes, Murrinh-patha verbs have 39 conjugation classes.      

\subsection{Tungusic} Tungusic languages are spoken principally in Russia, China and Mongolia. In Russia they are concentrated in north and eastern Siberia and in China in the east, in Manchuria. The largest languages in the family are Xibe, Evenki and Even; we use Evenki in the shared task. The languages are of the agglutinating morphological type with a moderate number of cases, 7 for Xibe and 13 for Evenki. In addition to case markers, Evenki marks  possession in nominals (including reflexive possession) and distinguishes between alienable and inalienable possession.  In terms of morphophonological processes, the languages exhibit vowel harmony, consonant alternations and phonological vowel length.
% \lb{noun-ordering}{\textit{stem - derivational affix(es) - alienable
%possession - number - case -  personal / %reflexive possession - clitic}}
% \lb{}{\textit{stem - derivational affix - evaluation - aspect - tense/mood - agreement}}

\subsection{Turkic} 
Languages of the Turkic family are primarily spoken in Central Asia. The family is morphologically concatenative, fusional, and suffixing. Turkic languages generally exhibit back vowel harmony, with the notable exception of Uzbek. In addition to harmony in backness, several languages also have labial vowel harmony (e.g., Kyrgyz, Turkmen, among others). In addition, most of the languages have dorsal consonant allophony that accompanies back vowel harmony. Additional morphophonological processes include vowel epenthesis and voicing assimilation. Selection of the inflectional allomorph can frequently be determined from the infinitive morpheme (which frequently reveals vowel backness and roundedness) and also the final segment of the stem. 
%\Eleanor{Check for Bashkir and others I didn't annotate.}

%% Eleanor
%\todo{should there be a footnote about how altaic is controversial?}
%\kat{Do you mean the definition of Altaic family? Probably yes }
%Francis

%%Kat
%  It should be ~1-3 paragraphs per language family. E.g, for Uralic languages, it might be worth mentioning spatial cases, how much their case systems differ from one another, how regular they are. Similarly, for conjugation: how large and regular paradigms are and how much they differ across languages (e.g, to what extent Finnic languages are similar to Permic).  
\subsection{Uralic} The Uralic languages are spoken in Russia from the north of Siberia to Scandinavia and Hungary in Europe. They are agglutinating with some subgroups displaying fusional characteristics (e.g., the Sámi languages). Many of the languages have vowel harmony. The languages have almost complete suffixal morphology and a medium-sized case inventory, ranging from 5--6 cases to numbers in the high teens. Many of the larger case paradigms are made up of spatial cases, sometimes with distinctions for direction and position.  Most of the languages have possessive suffixes, which can express possession, or agreement in non-finite clauses. The paradigms are largely regular, with few, if any, irregular forms. Many exhibit complex patterns of consonant gradation---consonant mutations that occur in specific morphological forms in some stems. Which gradation category a stem belongs to in often unpredictable.  The languages spoken in Russia are typically SOV, while those in Europe have SVO order.

\subsection{Uto-Aztecan}
The Uto-Aztecan family is represented by the Tohono O'odham (Papago--Pima) language spoken along the US--Mexico border in southern Arizona and northern Sonora. O'odham is agglutinative with a mixed prefixing and suffixing system. Nominal and verbal pluralization is frequently realized by partial reduplication of the initial consonant and/or vowel, and occasionally by final consonant deletion or null affixation. Processes targeting vowel length (shortening or lengthening) are also present. A small number of verbs exhibit suppletion in the past tense. 
%Eleanor(?)
\section{Data Preparation}

\subsection{Data Format}
Similar to previous years, training and development sets contain triples consisting of a lemma, a target form, and morphosyntactic descriptions (MSDs, or morphological tags).\footnote{Each MSD is a set of features separated by semicolons.} Test sets only contain two fields, i.e., target forms are omitted.
All data follows UTF-8 encoding. 

\subsection{Conversion and Canonicalization}
A significant amount of data for this task was extracted from corresponding (language-specific) grammars. In order to allow cross-lingual comparison, we manually converted their features (tags) into the UniMorph format \cite{sylak2016composition}.
We then canonicalized the converted language data\footnote{Using the UniMorph schema canonicalization script \url{https://github.com/unimorph/um-canonicalize}} to make sure all tags are consistently ordered and no category (e.g., ``Number'') is assigned two tags (e.g., singular and plural).\footnote{Conversion schemes and canonicalization scripts are available at \url{https://github.com/sigmorphon2020/task0-data}} 

\subsection{Splitting}

We use only noun, verb, and adjective forms to construct training, development, and evaluation sets. 
We de-duplicate annotations such that there are no multiple examples of exact lemma-form-tag matches. % nor ambiguous forms with the same lemma and tagset but different surface forms (rare).
\mans{Was this actually true for all languages' data sets?} 
\liz{not all had this (most didn't), but a number did have either this or duplicates of exact lemma-form-tag set and they could be frequent when so: including a couple from unimorph not just new data} 
\kat{I added Tables 6 and 7 in the Appendix with stats on inconsistencies}
\liz{awesome!!}
To create splits, we randomly sample 70\%, 10\%, and 20\% for train, development, and test, respectively. We cap the training set size to 100k examples for each language; where languages exceed this (e.g., Finnish), we subsample to this point, balancing lemmas such that all forms for a given lemma are either included or discarded. 
Some languages such as Zarma (dje), Tajik (tgk), Lingala (lin), Ludian* (lud), M\=aori (mao), Sotho (sot), V\~{o}ro (vro), Anglo-Norman (xno), and Zulu (zul) contain less than 400 training samples and are extremely low-resource.\footnote{We also note that Ludian contained inconsistencies in data due to merge of various dialects.}
\cref{tab:lang-stats} and \cref{tab:lang-stats-2} in the Appendix provide the number of samples for every language in each split, the number of samples per lemma, and statistics on inconsistencies in the data. 

\section{Baseline Systems}
The organizers provided two types of pre-trained baselines. Their use was optional.
\subsection{Non-neural}
The first baseline was a non-neural system that had been used as a baseline in earlier shared tasks on morphological reinflection \cite{cotterell-etal-2017-conll,cotterell2018conll}. The system first heuristically extracts lemma-to-form transformations; it assumes that these transformations are suffix- or prefix-based. A simple majority classifier is used to apply the most frequent suitable transformation to an input lemma, given the morphological tag, yielding the output form. See \citet{cotterell-etal-2017-conll} for further details.
\subsection{Neural}
Neural baselines were based on a neural transducer  \cite{wu2019exact}, which is essentially a hard monotonic attention model (\cd{mono-*}).
The second baseline is a transformer \cite{vaswani2017attention} adopted for character-level tasks that currently holds the state-of-the-art on the 2017 SIGMORPHON shared task data \cite[\md{trm-*}]{wu2020applying}.
Both models take the lemma and morphological tags as input and output the target inflection. The baseline is further expanded to include the data augmentation technique used by \citet[\md{-aug-}]{anastasopoulos2019pushing} (conceptually similar to the one proposed by \citet{silfverberg-etal-2017-data}). Relying on a simple character-level alignment between lemma and form, this technique replaces shared substrings of length $>3$ with random characters from the language's alphabet, producing hallucinated lemma--tag--form triples.
Both neural baselines were trained in mono- (\md{*-single}) and multilingual (shared parameters among the same family, \md{*-shared}) settings.

\section{Competing Systems}
%\Kat{Should be finalized once we get system description papers}
\begin{table*}[!ht]
    \centering
    \begin{adjustbox}{width=1\textwidth}
    \small
    \begin{tabular}{l|l|l|c|c|c|c}
    Team  &  Description & System &  \multicolumn{4}{c}{Model Features}\\
         & & & Neural & Ensemble & Multilingual & Hallucination\\
    \toprule
    \multirow{4}{*}{Baseline} & \multirow{4}{*}{\citet{wu2019exact}} & \md{mono-single}  & \checkmark &  &  &  \\
                                                        & &  \md{mono-aug-single} & \checkmark &  & & \checkmark\\
                                                        & &  \md{mono-shared} & \checkmark &  & \checkmark& \\
                                                        & &  \md{mono-aug-shared} & \checkmark &  & \checkmark & \checkmark\\
                                                        \cmidrule{2-7}
                            & \multirow{4}{*}{\citet{wu2020applying}}  &  \md{trm-single} & \checkmark &  & & \\
                                                        & &  \md{trm-aug-single} & \checkmark &  & & \checkmark\\
                                                         & &  \md{trm-shared} & \checkmark &  & \checkmark& \\
                                                       & &  \md{trm-aug-shared} & \checkmark &  & \checkmark & \checkmark \\
    \midrule
    \multirow{4}{*}{CMU Tartan} & \multirow{4}{*}{\citet{jayarao2020sigmorphon}} & \md{cmu\_tartan\_00-0}  & \checkmark& &  & \checkmark\\
                                                        & &  \md{cmu\_tartan\_00-1} & \checkmark &  & \checkmark & \checkmark\\
                                                         & &  \md{cmu\_tartan\_01-0} & \checkmark &  & & \checkmark\\
                                                         & &  \md{cmu\_tartan\_01-1} & \checkmark & & \checkmark& \checkmark\\
                                                         & &  \md{cmu\_tartan\_02-1} & \checkmark &  & \checkmark & \checkmark\\
    \midrule
    \multirow{2}{*}{CU7565} & \multirow{2}{*}{\citet{beemer2020sigmorphon}} & \md{CU7565-01-0} &   &  &  & \\
                                                         & &  \md{CU7565-02-0} &  &  & & \\
    \midrule
    CULing & \citet{liu2020sigmorphon} & \md{CULing-01-0}  & \checkmark & \checkmark  &  & \\
                                          
    \midrule
    \multirow{2}{*}{DeepSpin} & \multirow{2}{*}{\citet{peters2020sigmorphon}} & \md{deepspin-01-1}  & \checkmark &  & \checkmark & \\
                                                         & &  \md{deepspin-02-1} & \checkmark &  & \checkmark& \\                                                      
    \midrule
    \multirow{2}{*}{ETH Zurich} & \multirow{2}{*}{\citet{forster2020sigmorphon}} & \md{ETHZ00-1}  & \checkmark &  & \checkmark & \\
                                                         & &  \md{ETHZ02-1} & \checkmark & & \checkmark & \\
                                                         
    \midrule
    \multirow{2}{*}{Flexica} & \multirow{2}{*}{\citet{scherbakov2020sigmorphon}} & \md{flexica-01-0}  &  &  &  & \\
                                                     & &  \md{flexica-02-1} & \checkmark & &  \checkmark & \\         
                                                     & & \md{flexica-03-1} & \checkmark &  & \checkmark & \checkmark\\  
    \midrule
    IMS & \citet{yu2020sigmorphon} & \md{IMS-00-0}  & \checkmark & \checkmark &  & \checkmark\\
     \midrule
    LTI & \citet{murikinati2020sigmorphon} & \md{LTI-00-1}  & \checkmark  &  & \checkmark  & \checkmark \\
                               \midrule                        
  \multirow{4}{*}{NYU-CUBoulder} & \multirow{4}{*}{\citet{singer2020sigmorphon}} & \md{NYU-CUBoulder-01-0}  & \checkmark & \checkmark &  & \checkmark\\
                                                         & &  \md{NYU-CUBoulder-02-0} & \checkmark &  & & \checkmark\\
                                                         & &  \md{NYU-CUBoulder-03-0} & \checkmark & \checkmark & & \checkmark\\
                                                         & &  \md{NYU-CUBoulder-04-0} & \checkmark &  & & \checkmark\\
 \midrule
 UIUC & \citet{canby2020sigmorphon} & \md{uiuc-01-0}  & \checkmark &  &  & \\
    % \bottomrule
    \end{tabular}
    \end{adjustbox}
    \caption{The list of systems submitted to the shared task.}
    \label{tab:team_descr}
\end{table*}
As \cref{tab:team_descr} shows, 10 teams submitted 22 systems in total, out of which 19 were neural. 
Some teams such as \textbf{ETH Zurich} and \textbf{UIUC} built their models on top of the proposed baselines. In particular, \textbf{ETH Zurich} enriched each of the (multilingual) neural baseline models with exact decoding strategy that uses Dijkstra’s search algorithm.
\textbf{UIUC} enriched the transformer model with synchronous bidirectional decoding technique \cite{zhou2019synchronous} in order to condition the prediction of an affix character on its environment from both sides. (The authors demonstrate positive effects in Oto-Manguean, Turkic, and some Austronesian languages.)  

A few teams further improved models that were among top performers in previous shared tasks.
\textbf{IMS} and \textbf{Flexica} re-used the hard monotonic attention model from \cite{aharoni-goldberg-2017-morphological}. \textbf{IMS} developed an ensemble of two models (with left-to-right and right-to-left generation order) with a genetic algorithm for ensemble search \cite{haque2016heterogeneous} and iteratively provided hallucinated data.
\textbf{Flexica} submitted two neural systems. The first model (\md{flexica-02-1}) was multilingual (family-wise) hard monotonic attention model with improved alignment strategy. This model is further improved (\md{flexica-03-1}) by introducing a data hallucination technique which is based on phonotactic modelling of extremely low-resource languages \cite{shcherbakov2016phonotactic}.
\textbf{LTI} focused on their earlier model \cite{anastasopoulos2019pushing}, a neural multi-source encoder--decoder with two-step attention architecture, training it with hallucinated data, cross-lingual transfer, and romanization of scripts to improve performance on low-resource languages. 
\textbf{DeepSpin} reimplemented gated sparse two-headed attention model from \citet{peters2019sigmorphon} and trained it on all languages at once (massively multilingual). 
The team experimented with two modifications of the softmax function: sparsemax \cite[\md{deepspin-02-1}]{martins2016softmax} and 1.5-entmax \cite[\md{deepspin-01-1}]{peters2019sparse}.

Many teams based their models on the transformer architecture. \textbf{NYU-CUBoulder} experimented with a vanilla transformer model (\md{NYU-CUBoulder-04-0}), a pointer-generator transformer that allows for a copy mechanism (\md{NYU-CUBoulder-02-0}), and ensembles of three (\md{NYU-CUBoulder-01-0}) and five (\md{NYU-CUBoulder-03-0}) pointer-generator transformers. For languages with less than 1,000 training samples, they also generate hallucinated data.
\textbf{CULing} developed an ensemble of three (monolingual) transformers with identical architecture but different input data format. The first model was trained on the initial data format (lemma, target tags, target form). For the other two models the team used the idea of lexeme's principal parts \cite{finkel2007principal} and augmented the initial input (that only used the lemma as a source form) with entries corresponding to other (non-lemma) slots available for the lexeme. 
\kat{No mapping between system names and their description in the paper}
The \textbf{CMU Tartan} team compared performance of models with transformer-based and LSTM-based encoders and decoders. The team also compared monolingual to multilingual training in which they used several (related and unrelated) high-resource languages for low-resource language training.

Although the majority of submitted systems were neural, some teams experimented with non-neural approaches showing that in certain scenarios they might surpass neural systems.

\begin{table*}[!ht]
\begin{adjustbox}{width=1\textwidth}
    \centering
    \small
    \begin{tabular}{rl|rl|rl|rl|cc|c@{}c@{}c@{}c@{}}
    % \toprule
        \multicolumn{8}{c|}{Individual Language Rankings} &  \multicolumn{6}{c}{Final Ranking}\\
        \multicolumn{2}{c}{cly} & \multicolumn{2}{c}{ctp} &\multicolumn{2}{c}{czn} & \multicolumn{2}{c|}{zpv} &  & \multicolumn{1}{c}{avg} & \#1 \ & \#3 \ & \#4 \ & \#6 \\
    \toprule
    \md{uiuc}  & (1) &  \md{CULing} & (1) &  \md{deepspin} & (1) & \md{NYU-CUB} & (1) & \md{uiuc} & 1 & 4 \\
    \md{trm-single} & (1) &  \md{uiuc}  & (1) & \md{uiuc} & (1) & \md{CULing} & (1) & \md{trm-single} & 1 & 4 \\
    \md{CULing} & (3) & \md{trm-single} & (1) & \md{IMS} & (1) & \md{deepspin} & (1) & \md{CULing} & 1.5 & 3 & 1 \\
    \md{deepspin} & (3) & \md{IMS} & (4) & \md{NYU-CUB} & (1) & \md{uiuc}  & (1) & \md{deepspin} & 2.25 & 2 & 1 & 1\\
    \md{NYU-CUB} & (3) & \md{deepspin} & (4) & \md{CULing} & (1) & \md{trm-single} & (1) & \md{NYU-CUB} & 2.25 & 2 & 1 & 1 \\
    \md{IMS} & (6) & \md{NYU-CUB} & (4) & \md{trm-single} & (1) &  \md{IMS} & (1) & \md{IMS} & 3 & 2 & 0 & 1 & 1\\
    \bottomrule
    \end{tabular}
        \end{adjustbox}
    \caption{Illustration of our ranking method, over the four Zapotecan languages. Note: The final ranking is based on the actual counts (\#1,\#2, etc), not on the system's average rank.}
    \label{tab:rank_example}
\end{table*}

A large group of researchers from \textbf{CU7565} manually developed finite-state grammars for 25 languages (\md{CU7565-01-0}). They additionally developed a non-neural learner for all languages (\md{CU7565-02-0}) that uses hierarchical paradigm clustering (based on similarity of string transformation rules between inflectional slots). 
Another team, \textbf{Flexica}, proposed a model (\md{flexica-01-0}) conceptually similar to \citet{hulden2014semi}, although they did not attempt to reconstruct the paradigm itself and treated transformation rules independently assigning each of them a score based on its frequency and specificity as well as diversity of the characters surrounding the pattern.\footnote{English plural noun formation rule ``* $\rightarrow$ *s'' has high diversity whereas past tense rule such as ``*a* $\rightarrow$ *oo*'' as in \textit{(understand, understood)} has low diversity.}  
\begin{table}[H]
\centering
\begin{tabular}{c|c|c}
\toprule
System & Rank & Acc \\ \midrule
uiuc-01-0 & \textbf{2.4} & \textbf{90.5} \\ 
\rowcolor{LightGray!50} deepspin-02-1 & 2.9 & \textbf{90.9} \\ 
BASE: trm-single & 2.8 & 90.1 \\ 
\rowcolor{LightGray!50} CULing-01-0 & 3.2 & \textbf{91.2} \\ 
deepspin-01-1 & 3.8 & \textbf{90.5} \\ 
\rowcolor{LightGray!50} BASE: trm-aug-single & 3.7 & 90.3 \\ 
NYU-CUBoulder-04-0 & 7.1 & 88.8 \\ 
\rowcolor{LightGray!50} NYU-CUBoulder-03-0 & 8.9 & 88.8 \\ 
NYU-CUBoulder-02-0 & 8.9 & 88.7 \\ 
\rowcolor{LightGray!50} IMS-00-0 & 10.6 & 89.2 \\ 
NYU-CUBoulder-01-0 & 9.6 & 88.6 \\ 
\rowcolor{LightGray!50} BASE: trm-shared & 10.3 & 85.9 \\ 
BASE: mono-aug-single & 7.5 & 88.8 \\ 
\rowcolor{LightGray!50} cmu\_tartan\_00-0 & 8.7 & 87.1 \\ 
BASE: mono-single & 7.9 & 85.8 \\ 
\rowcolor{LightGray!50} cmu\_tartan\_01-1 & 9.0 & 87.1 \\ 
BASE: trm-aug-shared & 12.5 & 86.5 \\ 
\rowcolor{LightGray!50} BASE: mono-shared & 10.8 & 86.0 \\ 
cmu\_tartan\_00-1 & 9.4 & 86.5 \\ 
\rowcolor{LightGray!50} LTI-00-1 & 12.0 & 86.6 \\ 
BASE: mono-aug-shared & 12.8 & 86.8 \\ 
\rowcolor{LightGray!50} cmu\_tartan\_02-1 & 10.6 & 86.1 \\ 
cmu\_tartan\_01-0 & 10.9 & 86.6 \\ 
\rowcolor{LightGray!50} flexica-03-1 & 16.7 & 79.6 \\ 
ETHZ-00-1 & 20.1 & 75.6 \\ 
\rowcolor{LightGray!50} \textit{*CU7565-01-0} & \textit{24.1} & \textit{90.7} \\ 
flexica-02-1 & 17.1 & 78.5 \\ 
\rowcolor{LightGray!50} \textit{*CU7565-02-0} & \textit{19.2} & \textit{83.6} \\ 
ETHZ-02-1 & 17.0 & 80.9 \\ 
\rowcolor{LightGray!50} flexica-01-0 & 24.4 & 70.8 \\ 
\midrule
Oracle (Baselines) & & 96.1\\ 
Oracle (Submissions) & & 97.7\\
Oracle (All) & & 97.9\\
\bottomrule
\end{tabular}
\caption{Aggregate results on all languages. \textbf{Bolded} results are the ones which beat the best baseline. $*$ and \textit{italics} denote systems that did not submit outputs in all languages (their accuracy is a partial average).}
\label{tab:all}
\end{table}

\section{Evaluation}
This year, we instituted a slightly different evaluation regimen than in previous years, which takes into account the statistical significance of differences between systems and allows for an informed comparison across languages and families better than a simple macro-average.

The process works as follows:
\begin{enumerate}
    \item For each language, we rank the systems according to their accuracy (or Levenshtein distance). To do so, we use paired bootstrap resampling~\cite{koehn-2004-statistical}\footnote{We use 10,000 samples with 50\% ratio, and $p<0.005$.} to only take statistically significant differences into account. That way, any system which is the same (as assessed via statistical significance) as the best performing one is also ranked 1\textsuperscript{st} for that language.\mans{Is there a way to get a short description of how `statistical significance' is determined here. Not 100\% necessary but saves the reader a detour to Koehn (2004).}
    \item For the set of languages where we want collective results (e.g. languages within a linguistic genus), we aggregate the systems' ranks and re-rank them based on the amount of times they ranked 1\textsuperscript{st}, 2\textsuperscript{nd}, 3\textsuperscript{rd}, etc.
\end{enumerate}

Table~\ref{tab:rank_example} illustrates an example of this process using four Zapotecan languages and six systems.

\section{Results}
This year we had four winning systems (i.e., ones that outperform the best baseline): \md{CULing-01-0}, \md{deepspin-02-1}, \md{uiuc-01-0}, and \md{deepspin-01-1}, all neural. As \cref{tab:all} shows, they achieve over 90\% accuracy. Although \md{CULing-01-0} and \md{uiuc-01-0} are both monolingual transformers that do not use any hallucinated data, they follow different strategies to improve performance. The strategy proposed by \md{CULing-01-0} of enriching the input data with extra entries that included non-lemma forms and their tags as a source form, enabled their system to be among top performers on all language families; \md{uiuc-01-0}, on the other hand, did not modify the data but rather changed the decoder to be bidirectional and made family-wise fine-tuning of each (monolingual) model. The system is also among the top performers on all language families except Iranian. The third team, \textbf{DeepSpin}, trained and fine-tuned their models on all language data. Both models are ranked high (although the sparsemax model, \md{deepspin-02-1}, performs better overall) on most language groups with exception of Algic.  Sparsemax was also found useful by \textbf{CMU-Tartan}. 
The neural ensemble model with data augmentation from \textbf{IMS} team shows superior performance on languages with smaller data sizes (under 10,000 samples).
\textbf{LTI} and \textbf{Flexica} teams also observed positive effects of multilingual training and data hallucination on low-resource languages. The latter was also found useful in the ablation study made by \textbf{NYU-CUBoulder} team. 
Several teams aimed to address particular research questions; we will further summarize their results.

\begin{figure*}[h]
\includegraphics[width=\textwidth]{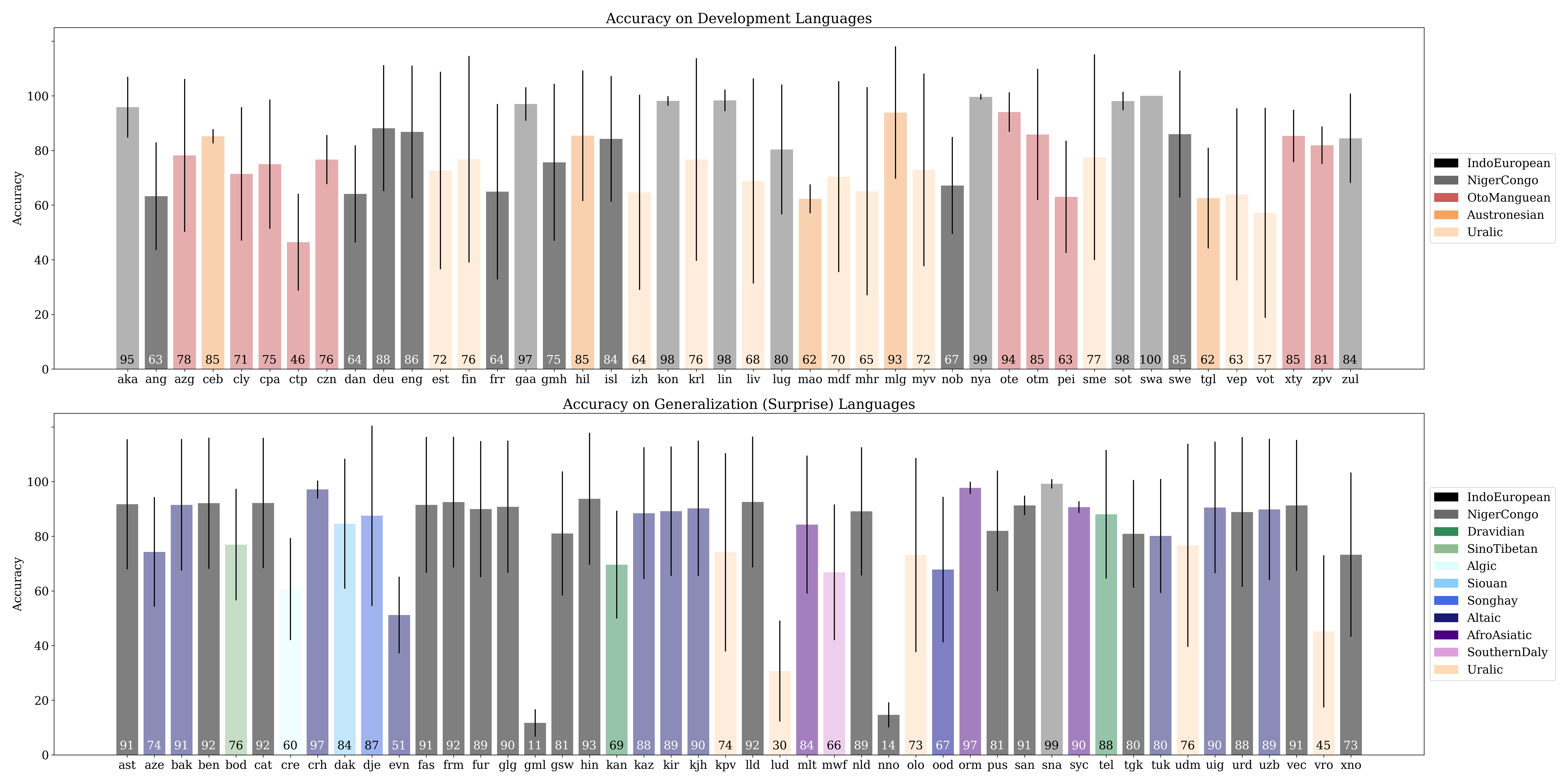}
\centering
\caption{Accuracy by language averaged across all the final submitted systems with their standard deviations. Language families are demarcated by color, with accuracy on development languages (top), and generalization languages (bottom).}
\label{fig:averageperformance}
\end{figure*}

\paragraph{Is developing morphological grammars manually worthwhile?}
This was the main question asked by \textbf{CU7565} who manually designed finite-state grammars for 25 languages. Paradigms of some languages were relatively easy to describe but neural networks also performed quite well on them even with a limited amount of data. For low-resource languages such as Ingrian and Tagalog the grammars demonstrate superior performance but this comes at the expense of a significant amount of person-hours.

%These results are further broken down by POS, and should help indicate what types of morphology systems find easy and hard.  They are in the files nounDifficulty.txt, verbDifficulty.txt, and adjDifficulty.txt.  I'm happy changing the definitions of hard and easy, or providing specific examples that fall in these categories.

\paragraph{What is the best training strategy for low-resource languages?}
Teams that generated hallucinated data highlighted its utility for low-resource languages. Augmenting the data with tuples where lemmas are replaced with non-lemma forms and their tags is another technique that was found useful. In addition, multilingual training and ensembles yield extra gain in terms of accuracy.
~\\

%\textbf{When does data hallucination help?}
\paragraph{Are the systems complementary?}
%\Kat{Added Oracle scores from Garrett (see error-analysis: oracle.scores). Shall we add family-/genuswise Oracle accuracy to the tables with results?}\Antonis{I can try to do that this weekend.}\kat{Thank you, that would be great!}
%\Garrett{from Garrett}
To address this question, we evaluate oracle scores for baseline systems, submitted systems, and all of them together.
Typically, as Tables 8--21 in the Appendix demonstrate, the baselines and the submissions are complementary - adding them together increases the oracle score.  Furthermore, while the full systems tend to dominate the partial systems (that were designed for a subset of languages, such as \md{CU7565-01-0}), there are a number of cases where the partial systems find the solution when the full systems don't - and these languages often then get even bigger gains when combined with the baselines.\christo{What do `full' and `partial' refer to here?}\kat{fixed}  This even happens when the accuracy of the baseline is very high - Finnish has baseline oracle of 99.89; full systems oracle of 99.91; submission oracle of 99.94 and complete oracle of 99.96, so an ensemble might be able to improve on the results. The largest gaps in oracle systems are observed in Algic, Oto-Manguean, Sino-Tibetan, Southern Daly, Tungusic, and Uto-Aztecan families.\footnote{Please see the results per language here:\\ \url{https://docs.google.com/spreadsheets/d/1ODFRnHuwN-mvGtzXA1sNdCi-jNqZjiE-i9jRxZCK0kg/edit?usp=sharing}}
\elenaklyachko{The link to the spreadsheet should be changed to read-only}\kat{Thank you! Fixed!}
~\\
\begin{figure*}[!h]
\includegraphics[width=\textwidth]{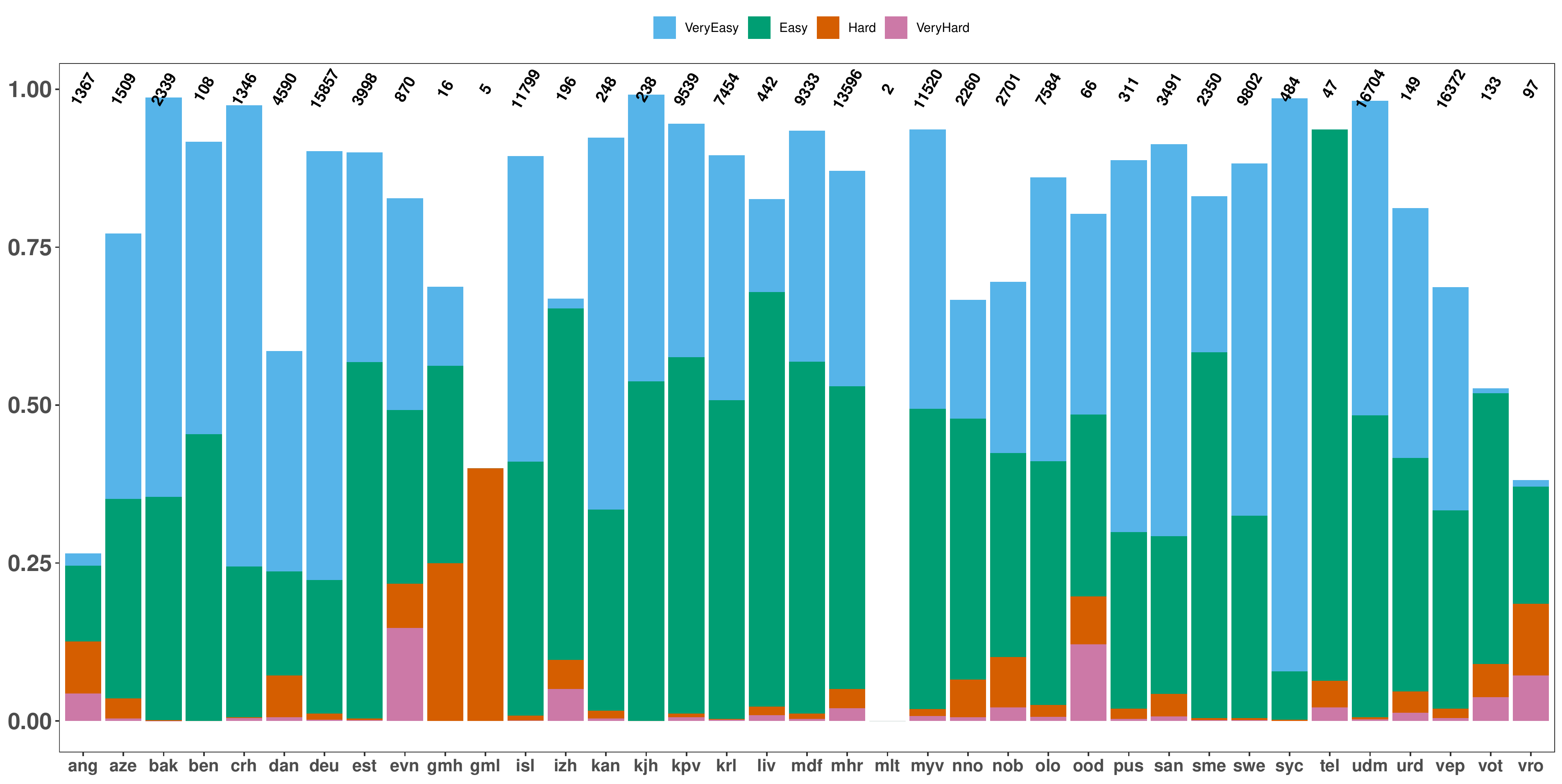}
\centering
\caption{Difficulty of Nouns: Percentage of test samples falling into each category. The total number of test samples for each language is outlined on the top of the plot.}
\label{fig:noun-diffic}
\end{figure*}
\begin{figure*}[!h]
\includegraphics[width=\textwidth]{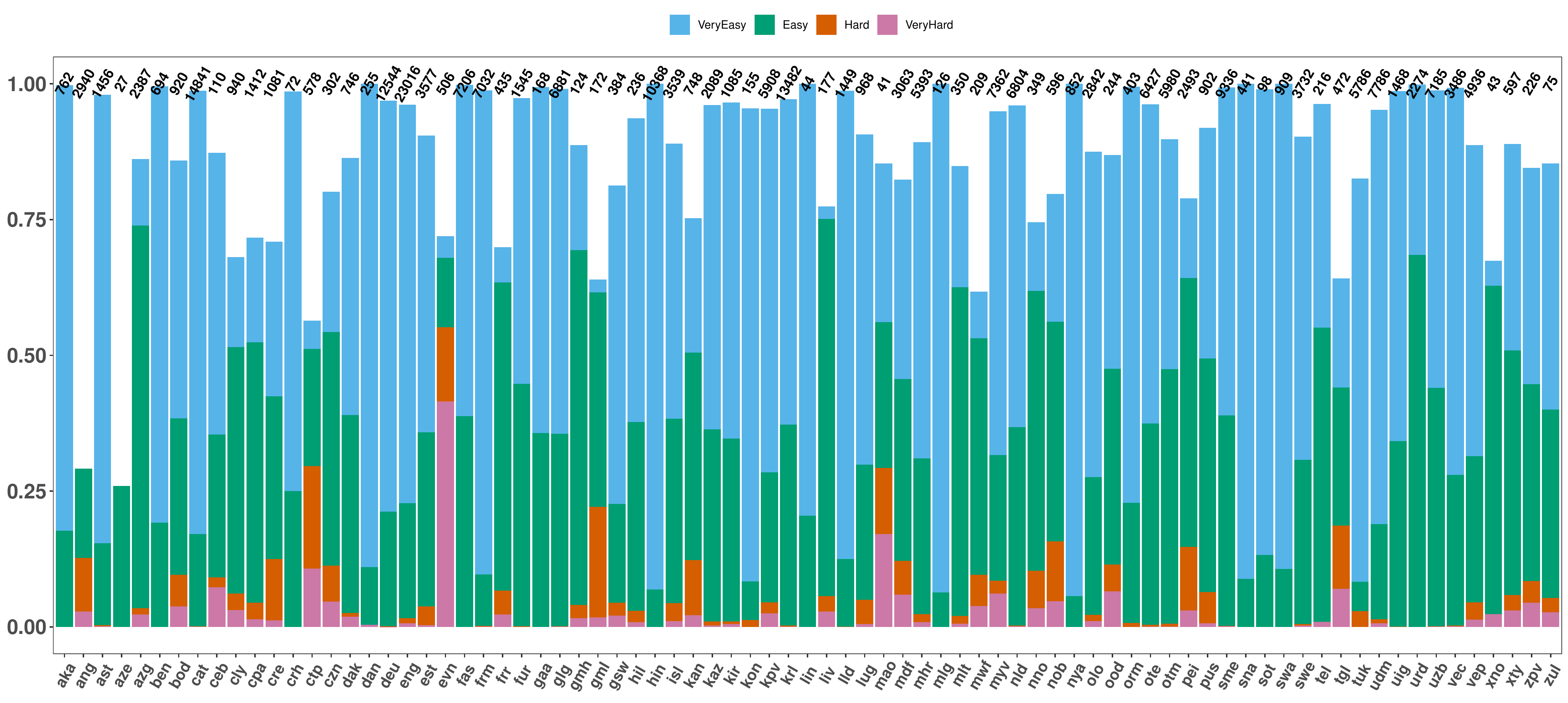}
\centering
\caption{Difficulty of Verbs: Percentage of test samples falling into each category. The total number of test samples for each language is outlined on the top of the plot.}
\label{fig:verb-diffic}
\end{figure*}

\begin{figure*}[!h]
\includegraphics[width=0.8\textwidth]{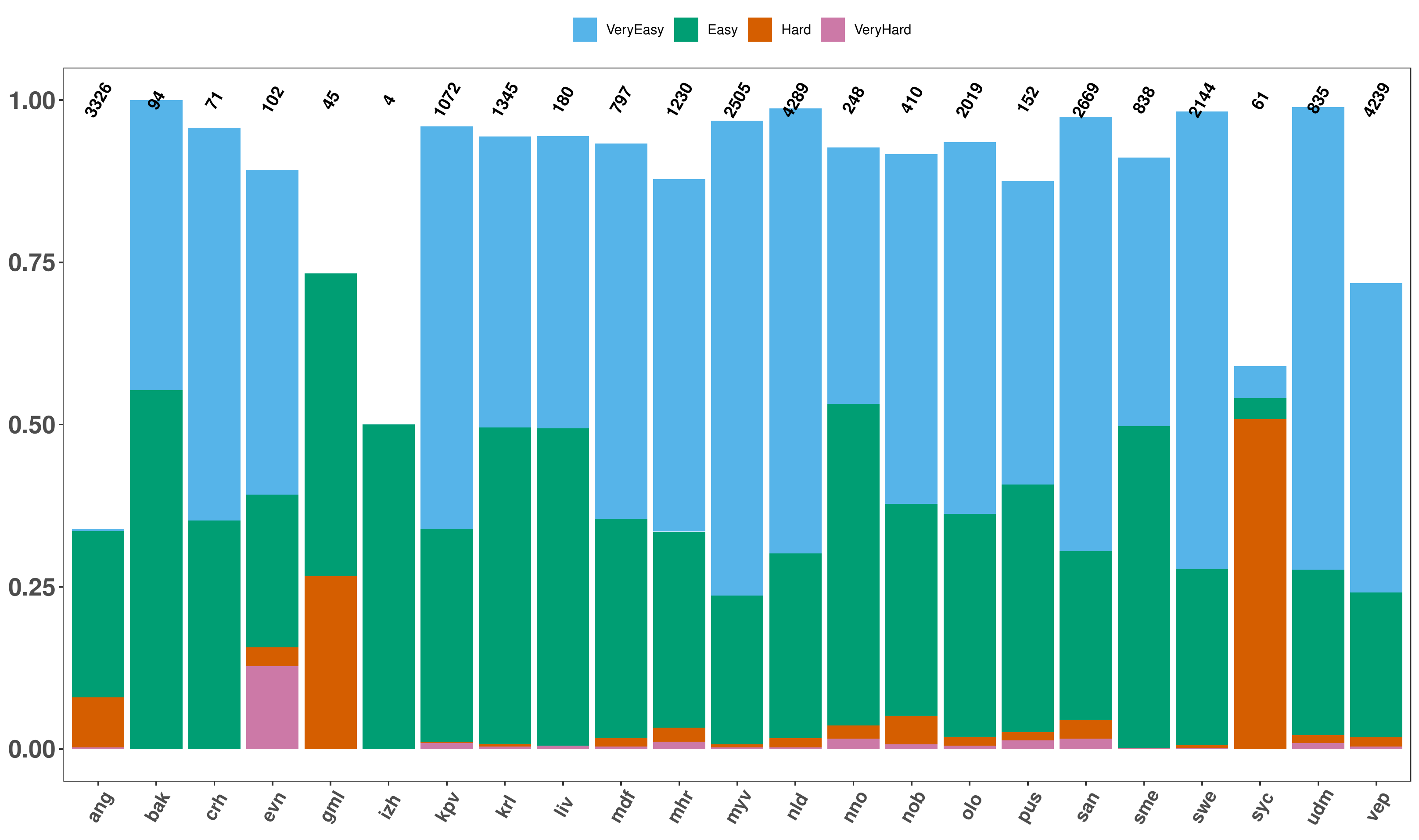}
\centering
\caption{Difficulty of Adjectives: Percentage of test samples falling into each category. The total number of test samples for each language is outlined on the top of the plot.}
\label{fig:adj-diffic}
\end{figure*}

\paragraph{Has morphological inflection become a solved problem in certain scenarios?}
%\kat{Not sure about this phrasing}\liz{`Has morphological inflection become a solved problem in certain scenarios?'}\kat{replaced! Thank you!}

The results shown in \cref{fig:averageperformance} suggest that for some of the development language families, such as Austronesian and Niger-Congo, the task was relatively easy, with most systems achieving high accuracy, whereas the task was more difficult for Uralic and Oto-Manguean languages, which showed greater variability in level of performance across submitted systems. 
Languages such as Ludic (lud), Norwegian Nynorsk (nno), Middle Low German (gml), Evenki (evn), and O'odham (ood) seem to be the most challenging languages based on simple accuracy.
For a more fine-grained study, we have classified test examples into four categories: ``very easy'', ``easy'', ``hard'', and ``very hard''.  
``Very easy'' examples are ones that all submitted systems got correct, while ``very hard'' examples are ones that no submitted system got correct.  ``Easy'' examples were predicted correctly for 80\% of systems, and ``hard'' were only correct in 20\% of systems. \cref{fig:noun-diffic}, \cref{fig:verb-diffic}, and \cref{fig:adj-diffic} represent percentage of noun, verb, and adjective samples that fall into each category and illustrate that most language samples are correctly predicted by majority of the systems.
For noun declension, Old English (ang), Middle Low German (gml), Evenki (evn), O'odham (ood), V\~{o}ro (vro) are the most difficult (some of this difficulty comes from language data inconsistency, as described in the following section). For adjective declension, Classic Syriac presents the highest difficulty (likely due to its limited data). \kat{Can someone comment on Syriac?}

\eleanor{is the y-axis on these figures accuracy? Or is difficulty a different metric?}
\kat{it's percentage of test samples that belong to the corresponding category (the plot is based on error-analysis/adjDifficulty.txt)}

\section{Error Analysis}
In our error analysis we follow the error type taxonomy proposed in 
\citet{gorman2019weird}. \eleanor{is this still true? does this just refer to the paragraph below? I have not been following a taxonomy...}
\kat{that's ok. It's very general: \%-ge of errors due to noise in the data; allomorphy;nonce words; orthography-related }
First, we evaluate systematic errors due to inconsistencies in the data, followed by an analysis of whether having seen the language or its family improved accuracy. We then proceed with an overview of accuracy for each of the language families. For a select number of families, we provide a more detailed analysis of the error patterns. 

\begin{figure*}[!h]
    \centering
    \includegraphics[width=\textwidth]{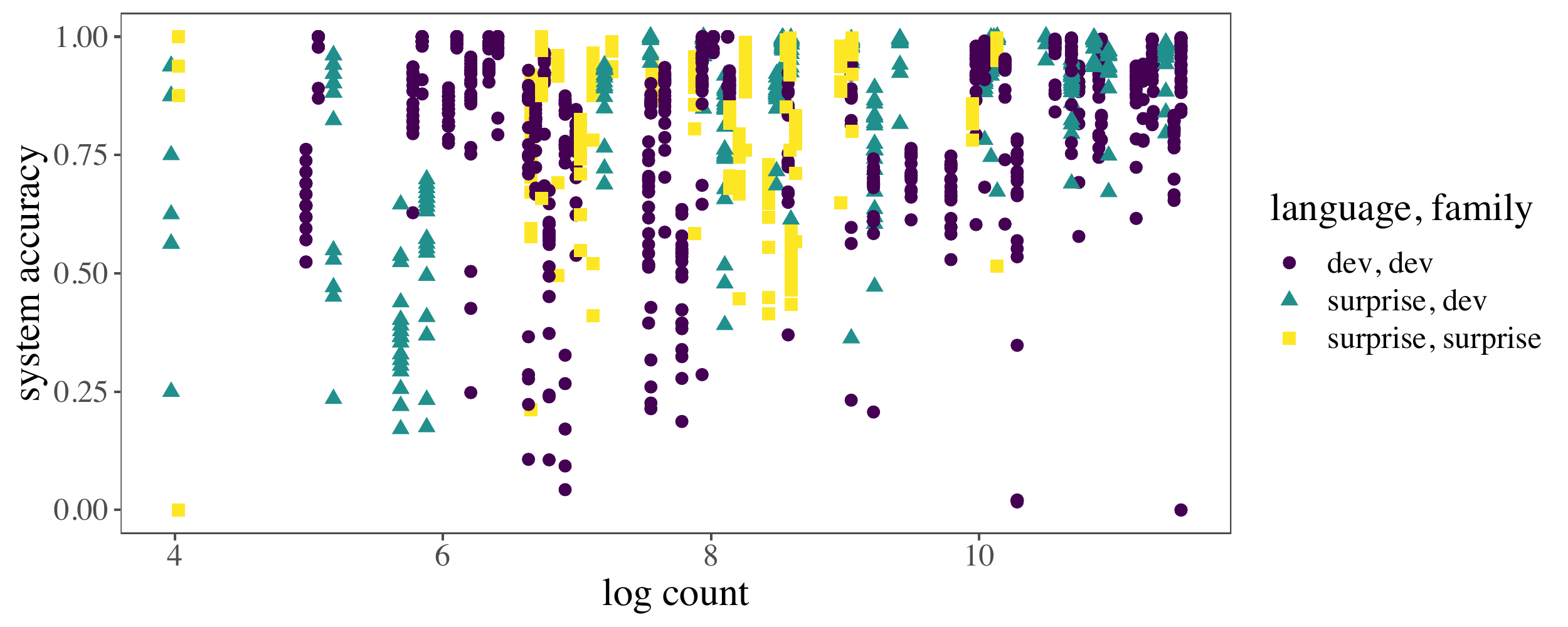}
    \caption{Accuracy for each system and language by the log size of the dataset. Points are color-coded according to language type: development language -- development family, surprise language -- development family, surprise language -- surprise family.}
    \label{fig:acc}
\end{figure*}

\cref{tab:lang-stats} and \cref{tab:lang-stats-2} provide the number of samples in the training, development, and test sets, percentage of inconsistent entries (the same lemma--tag pair has multiple infected forms) in them, percentage of contradicting entries (same lemma--tag pair occurring in train and development or test sets but assigned to different inflected forms), and percentage of entries in the development or test sets containing a lemma observed in the training set.
The train, development and test sets contain 2\%, 0.3\%, and 0.6\% inconsistent entries, respectively. Azerbaijani (aze), Old English (ang), Cree (cre), Danish (dan), Middle Low German (gml), Kannada (kan), Norwegian Bokm\r{a}l (nob), Chichimec (pei), and Veps (vep) had the highest rates of inconsistency.
These languages also exhibit the highest percentage of contradicting entries. The inconsistencies in some Finno-Ugric languages (such as Veps and Ludic) are due to dialectal variations. 

The overall accuracy of system and language pairings appeared to improve with an increase in the size of the dataset (\cref{fig:acc}; see also \cref{fig:sys-acc-fam} for accuracy trends by language family and \cref{fig:system-acc} for accuracy trends by system). Overall, the variance was considerable regardless of whether the language family or even the language itself had been observed during the Development Phase. A linear mixed-effects regression was used to assess variation in accuracy using fixed effects of language category, the size of the training dataset (log count), and their interactions, as well as random intercepts for system and language family accuracy.\footnote{Accuracy should ideally be assessed at the trial level using a logistic regression as opposed to a linear regression. By-trial accuracy was however not available at analysis time.} Language category was sum-coded with three levels: development language--development family, surprise language--development family, or surprise language--surprise family.

A significant effect of dataset size was observed, such that a one unit increase in log count corresponded to a 2\% increase in accuracy ($\beta$ = 0.019, $p < 0.001$). Language category type also significantly influenced accuracy: both development languages and surprise languages from development families were less accurate on average ($\beta_{dev-dev}$ = -0.145, $\beta_{sur-dev}$ = -0.167, each $p < 0.001$). These main effects were, however, significantly modulated by interactions with dataset size: on top of the main effect of dataset size, accuracy for development languages increased an additional $\approx$ 1.7\% ($\beta_{dev-dev \times size}$ = 0.017, $p < 0.001$) and accuracy for surprise languages from development families increased an additional $\approx$ 2.9\% ($\beta_{sur-dev \times size}$ = 0.029, $p < 0.001$). 

%Overall, the variance in per-language accuracy across systems was considerable regardless of whether the language family or even the language itself had been observed during the Development Phase. 

%\Eleanor{High variance in accuracy per language across systems regardless of whether the language's family or even the language itself had previously been observed. }

%\eleanor{I've added in these templatic paragraphs as fillers, but will certainly not be offended if someone goes in with the entropy analysis instead. I had the accuracies open in R so pulled the stats out, and didn't notice the entropy table until now. :S}

\paragraph{Afro-Asiatic:}
This family was represented by three languages. Mean accuracy across systems was above average at 91.7\%. Relative to other families, variance in accuracy was low, but nevertheless ranged from 41.1\% to 99.0\%.

\paragraph{Algic:}
This family was represented by one language, Cree. Mean accuracy across systems was below average at 65.1\%. Relative to other families, variance in accuracy was low, ranging from 41.5\% to 73\%. All systems appeared to struggle with the choice of preverbal auxiliary. Some auxiliaries were overloaded: `kitta' could refer to future, imperfective, or imperative. The morphological features for mood and tense were also frequently combined, such as SBJV+OPT (subjunctive plus optative mood). While the paradigms were very large, there were very few lemmas (28 impersonal verbs and 14 transitive verbs), which may have contributed to the lower accuracy. Interestingly, the inflections could largely be generated by rules.\footnote{Minor issues with the encoding of diacritics were identified, and will be corrected for release.}
%\eleanor{something got messed up in the character encoding in this dataset: there were apostrophes in the gold file, but I'm pretty sure I only used diacritics in the transcription. A lot of systems appeared to have difficulty with this decision between diacritics and apostrophes...}

\paragraph{Austronesian:}
This family was represented by five languages. Mean accuracy across systems was around average at 80.5\%. Relative to other families, variance in accuracy was high, with accuracy ranging from 39.5\% to 100\%.
One may notice a discrepancy among the difficulty in processing different Austronesian languages.
% This may be due to some languages in the family be much more irregular than others.
For instance, we see a difference of over 10\% in the baseline performance of Cebuano (84\%) and Hiligaynon (96\%).\footnote{We also note that some Hiligaynon entries contained multiple lemma forms (``bati/batian/pamatian'') for a single entry. We decided to leave it since we could not find any more information on which of the lemmas should be selected as the main. A similar issue was observed in Chichicapan Zapotec.}
% Cebuano only has six conjugations of each verb while Hiligaynon has 13, however, Cebuano contains many more irregularities \ran{I am pretty sure that this is true}.
This could come from the fact that Cebuano only has partial reduplication while Hiligaynon has full reduplication.
Furthermore, the prefix choice for Cebuano is more irregular, making it more difficult to predict the correct conjugation of the verb.

\paragraph{Dravidian:}
This family was represented by two languages: Kannada and Telugu. Mean accuracy across systems was around average at 82.2\%. Relative to other families, variance in accuracy was high: system accuracy ranged from 44.6\% to 96.0\%. Accuracy for Telugu was systematically higher than accuracy for Kannada.

\paragraph{Indo-European:}
This family was represented by 29 languages and four main branches. Mean accuracy across systems was slightly above average at 86.9\%. Relative to other families, variance in accuracy was very high: system accuracy ranged from 0.02\% to 100\%. For Indo-Aryan, mean accuracy was high (96.0\%) with low variance; for Germanic, mean accuracy was slightly below average (79.0\%) but with very high variance (ranging from 0.02\% to 99.5\%), for Romance, mean accuracy was high (93.4\%) but also had a high variance (ranging from 23.5\% to 99.8\%), and for Iranian, mean accuracy was high (89.2\%), but again with a high variance (ranging from 25.0\% to 100\%). Languages from the Germanic branch of the Indo-European family were included in the Development Phase.

\paragraph{Niger--Congo:}
This family was represented by ten languages. Mean accuracy across systems was very good at 96.4\%. Relative to other families, variance in accuracy was low, with accuracy ranging from 62.8\% to 100\%. Most languages in this family are considered low resource, and the resources used for data gathering may have been biased towards the languages' regular forms, as such this high accuracy may not be representative of the ``easiness'' of the task in this family. Languages from the Niger--Congo family was included in the Development Phase. 

\paragraph{Oto-Manguean:}
This family was represented by nine languages. Mean accuracy across systems was slightly below average at 78.5\%. Relative to other families, variance in accuracy was high, with accuracy ranging from 18.7\% to 99.1\%. Languages from the Oto-Manguean family were included in the Development Phase.

\paragraph{Sino-Tibetan:}
This family was represented by one language, Bodic. Mean accuracy across systems was average at 82.1\%, and variance across systems was also very low. Accuracy ranged from 67.9\% to 85.1\%.
The results are similar to those in \citet{di2019modelling} where majority of errors relate to allomorphy and impossible combinations of Tibetan unit components. 

\paragraph{Siouan:}
This family was represented by one language, Dakota. Mean accuracy across systems was above average at 89.4\%, and variance across systems was also low, despite the range from 0\% to 95.7\%. Dakota presented variable prefixing and infixing of person morphemes, along some complexities related to fortition processes. Determining the factor(s) that governed variation in affix position was difficult from a linguist's perspective, though many systems were largely successful. Success varied in the choice of the first or second person singular allomorphs which had increasing degrees of consonant strengthening (e.g., /wa/, /ma/, /mi/ /bde/, /bdu/ for the first person singular and /ya/, /na/, /ni/, /de/, or /du/ for the second person singular). In some cases, these fortition processes were overapplied, and in some cases, entirely missed. 

\paragraph{Songhay:}
This family was represented by one language, Zarma. Mean accuracy across systems was above average at 88.6\%, and variance across systems was relatively high. Accuracy ranged from 0\% to 100\%.

\paragraph{Southern Daly:}
This family was represented by one language, Murrinh-Patha. Mean accuracy across systems was below average at 73.2\%, and variance across systems was relatively high. Accuracy ranged from 21.2\% to 91.9\%.

\paragraph{Tungusic:}
This family was represented by one language, Evenki. The overall accuracy was the lowest across families. %\eleanor{help, are any of these genera that I'm referring to or are they all families?}
Mean accuracy was 53.8\% with very low variance across systems. Accuracy ranged from 43.5\% to 59.0\%.
\elenaklyachko{Added some analysis of the data}
The low accuracy is due to several factors. Firstly and primarily, the dataset was created from oral speech samples in various dialects of the language. The Evenki language is known to have rich dialectal variation. Moreover, there was little attempt at any standardization in the oral speech transcription. These peculiarities led to a high number of errors. For instance, some of the systems synthesized a wrong plural form for a noun ending in /-n/. Depending on the dialect, it can be /-r/ or /-l/, and there is a trend to have /-hVl/ for borrowed nouns. Deducing such a rule as well as the fact that the noun is a loanword is a hard task. Other suffixes may also have variable forms (such as /-kVllu/ vs /-kVldu/ depending on the dialect for the 2PL imperative. Some verbs have irregular past tense forms depending on the dialect and the meaning of the verb (e. g. /o:-/ 'to make' and 'to become'). Next, various dialects exhibit various vowel and consonant changes in suffixes. For example, some dialects (but not all of them) change /w/ to /b/ after /l/, and the systems sometimes synthesized a wrong form. The vowel harmony is complex: not all suffixes obey it, and it is also dialect-dependent. Some suffixes have variants (e. g., /-sin/ and /-s/ for SEMEL (semelfactive)), and the choice between them might be hard to understand. Finally, some of the mistakes are due to the markup scheme scarcity. For example, various past tense forms are all annotated as PST, or there are several comitative suffixes all annotated as COM. Moreover, some features are present in the word form but they receive no annotation at all. It is worth mentioning that some of the predictions could  theoretically be possible. To sum up, the Evenki case presents the challenges of oral non-standardized speech.

\paragraph{Turkic:}
This family was represented by nine languages. Mean accuracy across systems was relatively high at 93\%, and relative to other families, variance across systems was low. Accuracy ranged from 51.5\% to 100\%. Accuracy was lower for Azerbaijani and Turkmen, which after closer inspection revealed some slight contamination in the `gold' files. There was very marginal variation in the accuracy for these languages across systems. Besides these two, accuracies were predominantly above 98\%. A few systems struggled with the choice and inflection of the postverbal auxiliary in various languages (e.g., Kyrgyz, Kazakh, and Uzbek).

\paragraph{Uralic:}
This family was represented by 16 languages. Mean accuracy across systems was average at 81.5\%, but the variance across systems and languages was very high. Accuracy ranged from 0\% to 99.8\%. Languages from the Uralic family were included in the Development Phase. 

\paragraph{Uto-Aztecan:}
This family was represented by one language, O'odham. Mean accuracy across systems was slightly below average at 76.4\%, but the variance across systems and languages was fairly low. Accuracy ranged from 54.8\% to 82.5\%. The systems with higher accuracy may have benefited from better recall of suppletive forms relative to lower accuracy systems. 

%\Garrett{There are 2 entropy tables - one with entropy of errors across systems, and one of entropy of predictions across systems (ie, errors + correct predictions) - I've added all these numbers (entropy and oracle scores) into the results spreadsheet, to make them a little easier to use.}
%\Garrett{toward the goal of number 2 (error overlap), I've calculated the average number of different errors made among the systems, as well as the error entropy.  I also calculated the prediction entropy (same as error entropy, but also include systems that got the answer right).  These values should help us determine which languages were particularly variable when it came to system output, which I think might guide us in our error analysis.

%Another note: while calculating the oracle scores, it came to my attention that a very small number of errors were due to participants changing the format of the MSDs (ie, V.PTCP to V;PTCP, or PST+IMMED to PST;IMMED.  It's likely an insignificant number, but might be worth a footnote. }

\section{Conclusion}

This years's shared task on morphological reinflection focused on building models that could generalize across an extremely typologically diverse set of languages, many from understudied language families and with limited available text resources. As in previous years, neural models performed well, even in relatively low-resource cases. 
Submissions were able to make productive use of multilingual training to take advantage of commonalities across languages in the dataset. 
Data augmentation techniques such as hallucination helped fill in the gaps and allowed networks to generalize to unseen inputs. 
These techniques, combined with architecture tweaks like sparsemax, resulted in excellent overall performance on many languages (over 90\% accuracy on average). 
However, the task's focus on typological diversity revealed that some morphology types and language families (Tungusic, Oto-Manguean, Southern Daly) remain a challenge for even the best systems. 
These families are extremely low-resource, represented in this dataset by few or a single language. 
This makes cross-linguistic transfer of similarities by multilanguage training less viable.
They may also have morphological properties and rules (e.g., Evenki is agglutinating with many possible forms for each lemma) that are particularly difficult for machine learners to induce automatically from sparse data. 
For some languages (Ingrian, Tajik, Tagalog, Zarma, and Lingala), optimal performance was only achieved in this shared task by hand-encoding linguist knowledge in finite state grammars.
It is up to future research to imbue models with the right kinds of linguistic inductive biases to overcome these challenges.\christo{This is a stub. Feel free to change (e.g. add explicit list of winners?}

\section*{Acknowledgements}
We would like to thank each of the participants for their time and effort in developing their task systems. We also thank Jason Eisner for organization and guidance. We thank Vitalij Chernyavskij for his help with V\~{o}ro and Umida Boltaeva and Bahriddin Abdiev for their contribution in Uzbek data annotation.

\bibliography{anthology,acl2020}
\bibliographystyle{acl_natbib}

\appendix
\onecolumn
\section{Language data statistics}
\begin{table}[!h]
\begin{adjustbox}{width=1\textwidth}
\small
\centering
\small
\begin{tabular}{c|r|r|r|r|r|r|>{\hspace{2em}}r|r|>{\hspace{2em}}r|r}
\toprule
\multicolumn{1}{c}{\textbf{Lang}}&\multicolumn{3}{c}{\textbf{Total}}&\multicolumn{3}{c}{\textbf{Inconsistency} (\%)}&\multicolumn{2}{c}{\textbf{Contradiction} (\%)}&\multicolumn{2}{c}{\textbf{In Vocabulary} (\%)}\\
\cmidrule(lr){1-1} \cmidrule(lr){2-4} \cmidrule(lr){5-7} \cmidrule(lr){8-9} \cmidrule(lr){10-11}
 &Train& Dev & Test &Train&Dev&Test&Dev&Test&Dev&Test\\
\midrule
aka&2793&380&763&0.0&0.0&0.0&0.0&0.0&24.7&12.5\\
ang&29270&4122&8197&11.8&1.8&3.4&21.6&21.9&35.1&21.3\\
ast&5096&728&1457&0.0&0.0&0.0&0.0&0.0&23.9&12.4\\
aze&5602&801&1601&11.9&1.9&4.0&22.3&20.9&31.5&20.2\\
azg&8482&1188&2396&0.8&0.0&0.0&1.3&1.1&26.9&13.8\\
bak&8517&1217&2434&0.0&0.0&0.0&0.0&0.0&59.8&40.1\\
ben&2816&402&805&0.0&0.0&0.0&0.0&0.0&29.9&16.0\\
bod&3428&466&936&1.0&0.2&0.3&2.4&1.9&80.0&73.4\\
cat&51944&7421&14842&0.0&0.0&0.0&0.0&0.0&20.8&10.4\\
ceb&420&58&111&1.0&0.0&0.0&0.0&2.7&72.4&62.2\\
cly&3301&471&944&0.0&0.0&0.0&0.0&0.0&37.4&19.3\\
cpa&5298&727&1431&3.4&0.6&0.8&6.6&4.3&60.2&39.8\\
cre&4571&584&1174&18.5&2.1&4.9&29.8&29.6&5.5&2.7\\
crh&5215&745&1490&0.0&0.0&0.0&0.0&0.0&77.4&60.7\\
ctp&2397&313&598&15.9&1.6&3.0&22.0&21.7&52.7&34.1\\
czn&1088&154&305&0.2&0.0&0.0&1.3&0.0&86.4&74.8\\
dak&2636&376&750&0.0&0.0&0.0&0.0&0.0&75.5&55.7\\
dan&17852&2550&5101&16.5&2.5&5.0&34.5&32.9&71.4&51.8\\
deu&99405&14201&28402&0.0&0.0&0.0&0.0&0.0&55.8&37.8\\
dje&56&9&16&0.0&0.0&0.0&0.0&0.0&100.0&87.5\\
eng&80865&11553&23105&1.1&0.2&0.4&2.1&1.9&80.3&66.2\\
est&26728&3820&7637&2.7&0.4&0.8&6.1&5.1&22.4&11.6\\
evn&5413&774&1547&9.6&2.8&4.3&8.9&10.0&38.9&32.5\\
fas&25225&3603&7208&0.0&0.0&0.0&0.0&0.0&7.6&3.8\\
fin&99403&14201&28401&0.0&0.0&0.0&0.0&0.0&32.6&17.2\\
frm&24612&3516&7033&0.0&0.0&0.0&0.0&0.0&17.1&8.6\\
frr&1902&224&477&4.0&0.0&1.7&9.8&6.1&22.8&10.7\\
fur&5408&772&1546&0.0&0.0&0.0&0.0&0.0&21.6&10.9\\
gaa&607&79&169&0.0&0.0&0.0&0.0&0.0&74.7&47.3\\
glg&24087&3441&6882&0.0&0.0&0.0&0.0&0.0&14.1&7.1\\
gmh&496&71&141&1.2&0.0&0.0&5.6&2.8&38.0&20.6\\
gml&890&127&255&17.3&3.1&5.5&22.8&27.8&39.4&20.4\\
gsw&1345&192&385&0.0&0.0&0.0&0.0&0.0&55.7&35.6\\
hil&859&116&238&0.0&0.0&0.0&0.0&0.0&59.5&36.6\\
hin&36300&5186&10372&0.0&0.0&0.0&0.0&0.0&5.0&2.5\\
isl&53841&7690&15384&1.0&0.1&0.3&1.9&2.0&48.8&29.5\\
izh&763&112&224&0.0&0.0&0.0&0.0&0.0&42.9&22.3\\
kan&3670&524&1049&13.2&2.7&4.7&18.7&20.7&21.9&14.0\\
kaz&7852&1063&2113&1.1&0.2&0.4&1.9&1.8&10.6&5.3\\
kir&3855&547&1089&0.0&0.0&0.0&0.0&0.0&17.9&9.0\\
kjh&840&120&240&0.0&0.0&0.0&0.0&0.0&50.8&30.4\\
kon&568&76&156&0.0&0.0&0.0&0.0&0.0&78.9&71.8\\
kpv&57919&8263&16526&0.0&0.0&0.0&0.0&0.0&48.8&35.0\\
krl&80216&11225&22290&0.2&0.0&0.0&0.3&0.3&19.7&10.3\\
lin&159&23&46&0.0&0.0&0.0&0.0&0.0&100.0&73.9\\
liv&2787&398&802&0.0&0.0&0.0&0.0&0.0&40.7&24.1\\
\bottomrule
\end{tabular}
\end{adjustbox}
\caption{Number of samples in training, development, test sets, as well as statistics on systematic errors (inconsistency) and percentage of samples with lemmata observed in the training set.}
\label{tab:lang-stats}
\end{table}

\begin{table}[!h]
\begin{adjustbox}{width=1\textwidth}
\small
\centering
\small
\begin{tabular}{c|r|r|r|r|r|r|>{\hspace{2em}}r|r|>{\hspace{2em}}r|r}
\toprule
\multicolumn{1}{c}{\textbf{Lang}}&\multicolumn{3}{c}{\textbf{Total}}&\multicolumn{3}{c}{\textbf{Inconsistency} (\%)}&\multicolumn{2}{c}{\textbf{Contradiction} (\%)}&\multicolumn{2}{c}{\textbf{In Vocabulary} (\%)}\\
\cmidrule(lr){1-1} \cmidrule(lr){2-4} \cmidrule(lr){5-7} \cmidrule(lr){8-9} \cmidrule(lr){10-11}
 &Train& Dev & Test &Train&Dev&Test&Dev&Test&Dev&Test\\
\midrule
lld&5073&725&1450&0.0&0.0&0.0&0.0&0.0&24.3&12.3\\
lud&294&41&82&7.8&0.0&3.7&9.8&11.0&31.7&20.7\\
lug&3420&489&977&4.0&0.6&0.8&5.1&7.6&18.2&9.1\\
mao&145&21&42&0.0&0.0&0.0&0.0&0.0&61.9&81.0\\
mdf&46362&6633&13255&1.6&0.2&0.5&3.1&3.3&49.0&35.1\\
mhr&71143&10081&20233&0.3&0.0&0.0&0.4&0.5&48.8&34.3\\
mlg&447&62&127&0.0&0.0&0.0&0.0&0.0&90.3&74.0\\
mlt&1233&176&353&0.1&0.0&0.0&0.6&0.0&52.3&30.6\\
mwf&777&111&222&2.6&0.0&0.9&2.7&4.5&25.2&13.1\\
myv&74928&10738&21498&1.7&0.3&0.5&3.1&3.1&45.5&32.7\\
nld&38826&5547&11094&0.0&0.0&0.0&0.0&0.0&58.2&38.4\\
nno&10101&1443&2887&3.4&0.4&1.0&6.0&6.8&80.0&70.2\\
nob&13263&1929&3830&10.5&1.8&3.1&18.5&19.7&80.5&70.5\\
nya&3031&429&853&0.0&0.0&0.0&0.0&0.0&46.4&26.5\\
olo&43936&6260&12515&1.4&0.3&0.5&3.3&2.9&83.0&70.8\\
ood&1123&160&314&0.4&0.0&0.0&1.9&1.0&70.0&58.0\\
orm&1424&203&405&0.2&0.0&0.2&0.5&0.7&41.9&22.7\\
ote&22962&3231&6437&0.4&0.1&0.1&0.5&0.8&48.4&29.5\\
otm&21533&3020&5997&0.9&0.1&0.3&1.8&1.7&49.4&29.4\\
pei&10017&1349&2636&15.8&2.6&4.9&21.5&21.4&9.1&4.7\\
pus&4861&695&1389&3.9&0.6&1.6&9.9&7.7&34.2&23.0\\
san&22968&3188&6272&3.1&0.5&0.9&4.5&5.5&26.9&14.6\\
sme&43877&6273&12527&0.0&0.0&0.0&0.0&0.0&28.2&16.3\\
sna&1897&246&456&0.0&0.0&0.0&0.0&0.0&31.3&18.0\\
sot&345&50&99&0.0&0.0&0.0&0.0&0.0&48.0&25.3\\
swa&3374&469&910&0.0&0.0&0.0&0.0&0.0&20.7&10.5\\
swe&54888&7840&15683&0.0&0.0&0.0&0.0&0.0&70.6&51.9\\
syc&1917&275&548&3.5&1.5&0.4&7.6&8.6&47.3&28.1\\
tel&952&136&273&1.4&0.0&1.1&0.7&2.6&62.5&39.6\\
tgk&53&8&16&0.0&0.0&0.0&0.0&0.0&0.0&0.0\\
tgl&1870&236&478&7.6&1.3&1.0&11.9&10.0&74.2&55.6\\
tuk&20963&2992&5979&9.5&1.5&3.2&16.8&16.0&16.7&8.3\\
udm&88774&12665&25333&0.0&0.0&0.0&0.0&0.0&38.1&24.8\\
uig&5372&750&1476&0.3&0.0&0.0&0.3&0.5&12.0&6.1\\
urd&8486&1213&2425&0.0&0.0&0.0&0.0&0.0&9.4&6.0\\
uzb&25199&3596&7191&0.0&0.0&0.0&0.0&0.0&11.9&6.0\\
vec&12203&1743&3487&0.0&0.0&0.0&0.0&0.0&20.8&10.6\\
vep&94395&13320&26422&10.9&1.8&3.3&19.3&19.8&25.1&12.9\\
vot&1003&146&281&0.0&0.0&0.0&0.0&0.0&35.6&19.6\\
vro&357&51&103&1.1&0.0&0.0&2.0&1.0&70.6&50.5\\
xno&178&26&51&0.0&0.0&0.0&0.0&0.0&19.2&9.8\\
xty&2110&299&600&0.1&0.3&0.0&0.3&1.3&78.6&65.8\\
zpv&805&113&228&0.0&0.0&0.4&2.7&0.9&78.8&78.9\\
zul&322&42&78&1.9&0.0&0.0&2.4&0.0&83.3&66.7\\
\midrule
\bf TOTAL&1574004&223649&446580&2.0&0.3&0.6&3.6&3.6&41.1&27.9\\
\bottomrule
\end{tabular}
\end{adjustbox}
\caption{Number of samples in training, development, test sets, as well as statistics on systematic errors (inconsistency) and percentage of samples with lemmata observed in the training set.}
\label{tab:lang-stats-2}
\end{table}

\newpage\clearpage
\section{Accuracy trends}
\begin{figure}[!ht]
    \centering
    \includegraphics[width=\textwidth]{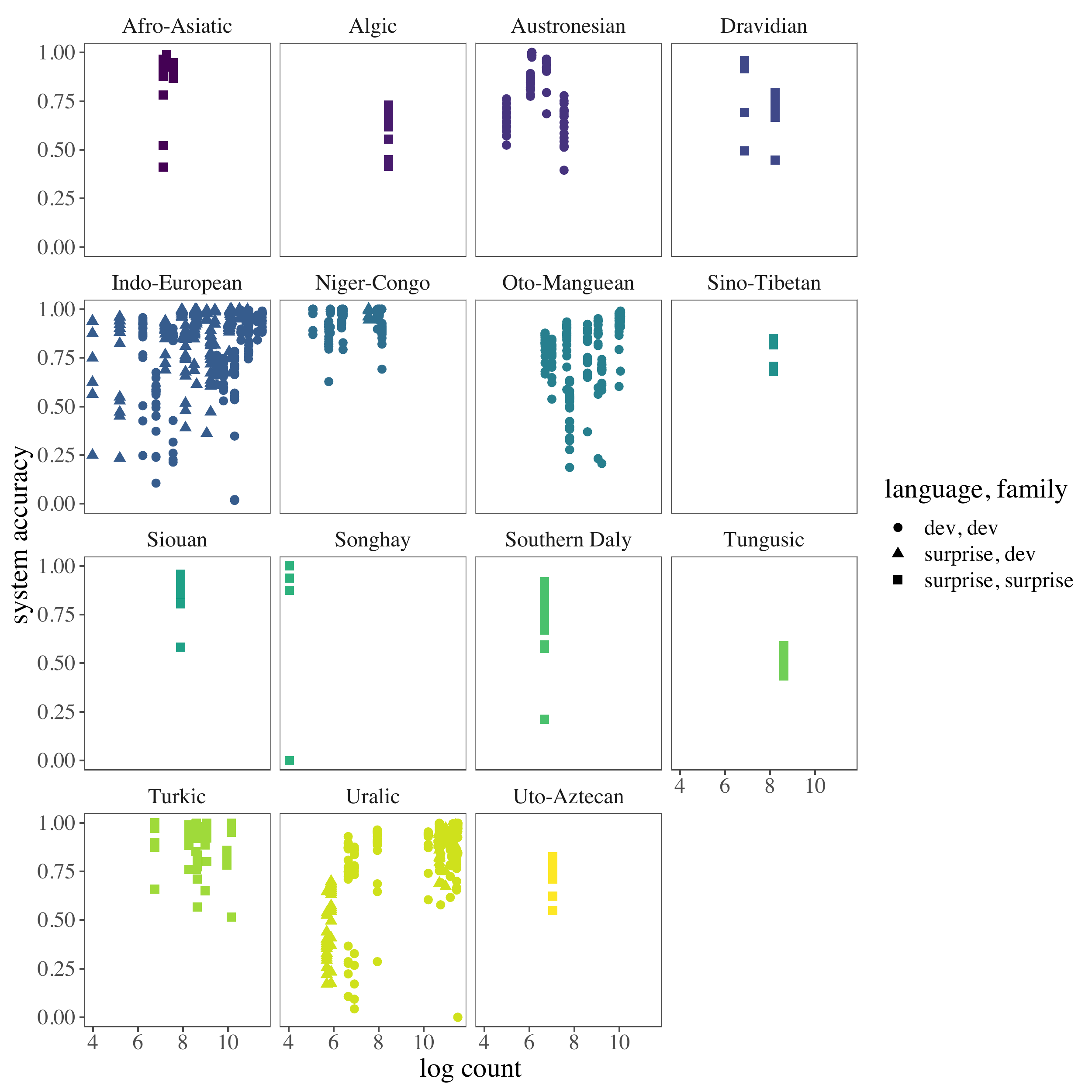}
    \caption{Accuracy for each system and language by the log size of the dataset, \textbf{grouped by language family}. Points are color-coded according to language family, and shape-coded according to language type: development language -- development family, surprise language -- development family, surprise language -- surprise family.}
    \label{fig:sys-acc-fam}
\end{figure}

\newpage
\begin{figure}[!ht]
    \centering
    \includegraphics[width=\textwidth]{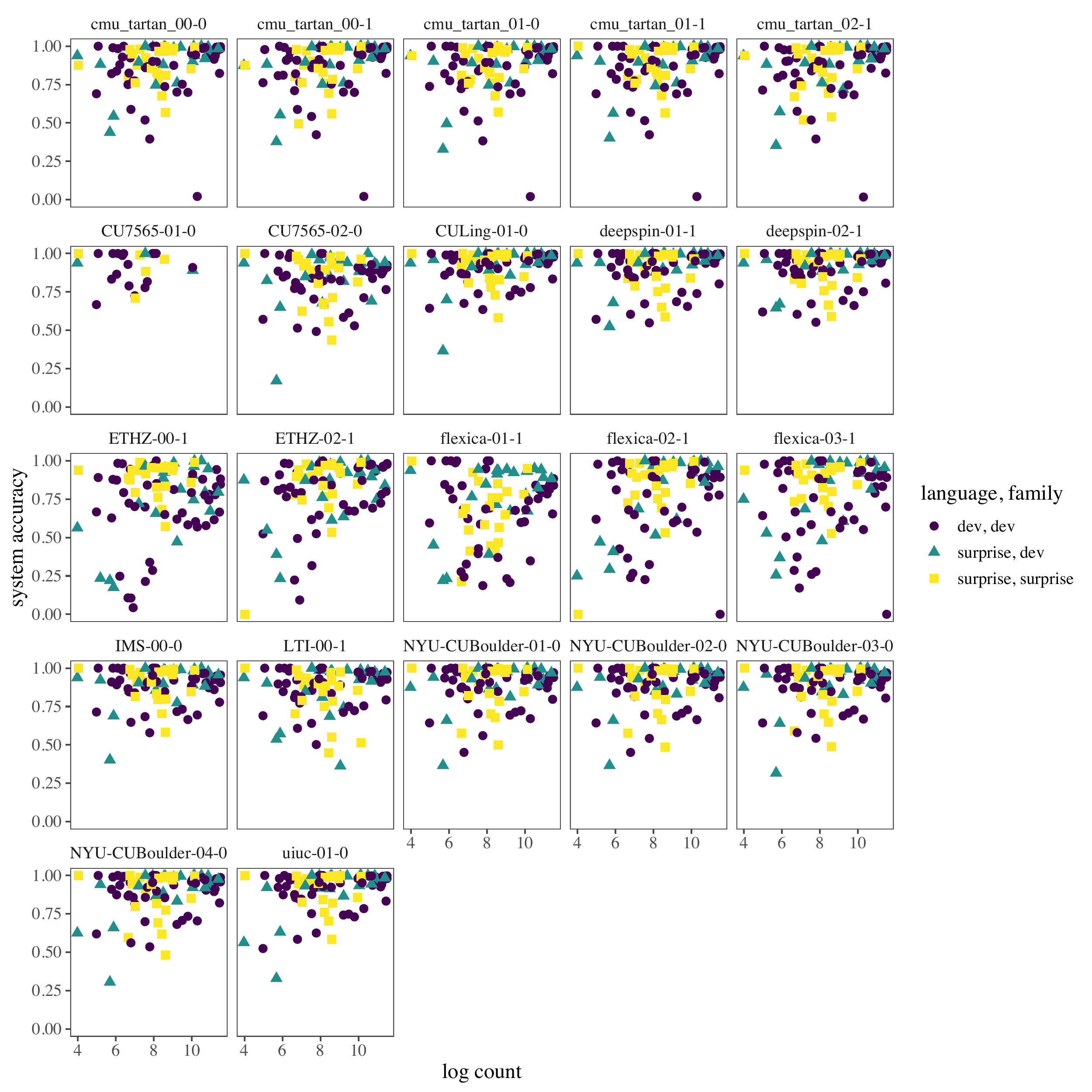}
    \caption{Accuracy for each language by the log size of the dataset, \textbf{grouped by submitted system}. Points are color- and shape-coded according to language type: development language -- development family, surprise language -- development family, surprise language -- surprise family.}
    \label{fig:system-acc}
\end{figure}

%\section{Results per Language Family}
\begin{table}[!htb]
\caption{\textbf{Results per Language Family: Afro-Asiatic and Algic}}
\centering
 \begin{subtable}{.5\linewidth}
 \centering
 %\begin{table}
%\centering
\begin{tabular}{c|c|c}
\toprule
System & Rank & Acc \\ \midrule
uiuc-01-0 & 1.0 & 96.4 \\ 
CULing-01-0 & 1.0 & 96.3 \\ 
\rowcolor{LightGray} deepspin-02-1 & 3.7 & 95.2 \\ 
BASE: trm-single & 4.0 & 95.5 \\ 
BASE: trm-aug-single & 4.0 & 95.0 \\ 
deepspin-01-1 & 4.0 & 94.7 \\ 
NYU-CUBoulder-01-0 & 4.0 & 94.4 \\ 
NYU-CUBoulder-02-0 & 4.0 & 94.4 \\ 
\rowcolor{LightGray} NYU-CUBoulder-04-0 & 9.7 & 94.3 \\ 
BASE: mono-single & 6.3 & 92.8 \\ 
cmu\_tartan\_00-0 & 6.3 & 92.7 \\ 
\rowcolor{LightGray} cmu\_tartan\_01-0 & 9.3 & 89.6 \\ 
\rowcolor{LightGray} cmu\_tartan\_01-1 & 9.3 & 89.4 \\ 
cmu\_tartan\_02-1 & 10.0 & 80.9 \\ 
\rowcolor{LightGray} ETHZ-00-1 & 6.7 & 94.7 \\ 
\rowcolor{LightGray} BASE: trm-shared & 6.7 & 94.2 \\ 
\rowcolor{LightGray} BASE: trm-aug-shared & 6.7 & 94.0 \\ 
\rowcolor{LightGray} IMS-00-0 & 6.7 & 93.6 \\ 
\rowcolor{LightGray} BASE: mono-aug-single & 6.7 & 93.5 \\ 
NYU-CUBoulder-03-0 & 12.3 & 93.7 \\ 
\rowcolor{LightGray} flexica-02-1 & 9.3 & 92.9 \\ 
\rowcolor{LightGray} ETHZ-02-1 & 9.3 & 92.3 \\ 
\rowcolor{LightGray} flexica-03-1 & 9.3 & 92.1 \\ 
\rowcolor{LightGray} BASE: mono-shared & 9.3 & 91.5 \\ 
\textit{*CU7565-01-0} & \textit{19.3} & \textit{93.7} \\ 
\rowcolor{LightGray} BASE: mono-aug-shared & 16.0 & 89.8 \\ 
CU7565-02-0 & 15.0 & 91.6 \\ 
\rowcolor{LightGray} cmu\_tartan\_00-1 & 17.7 & 91.7 \\ 
\rowcolor{LightGray} LTI-00-1 & 17.7 & 91.3 \\ 
flexica-01-1 & 28.3 & 73.4 \\ 
\midrule
Oracle (Baselines) & & 98.7 \\
Oracle (Submissions) & & 99.7 \\
Oracle (All) & & 99.8 \\
\bottomrule
\end{tabular}
%\caption{Results on the Afro-Asiatic genus (3 languages).}
%\label{tab:AfroAsiatic}
%\end{table}

 \caption{Results on the Afro-Asiatic family (3 languages)}
 \end{subtable}%
 \begin{subtable}{.5\linewidth}
  \centering
 %\begin{table}
%\centering
\begin{tabular}{c|c|c}
\toprule
System & Rank & Acc \\ \midrule
CULing-01-0 & 1.0 & 73.0 \\ 
flexica-03-1 & 1.0 & 70.4 \\ 
IMS-00-0 & 1.0 & 70.3 \\ 
uiuc-01-0 & 1.0 & 70.3 \\ 
ETHZ-02-1 & 1.0 & 69.4 \\ 
cmu\_tartan\_02-1 & 1.0 & 69.4 \\ 
flexica-02-1 & 1.0 & 69.4 \\ 
\rowcolor{LightGray} cmu\_tartan\_00-1 & 8.0 & 69.2 \\ 
\rowcolor{LightGray} BASE: mono-aug-shared & 8.0 & 68.5 \\ 
\rowcolor{LightGray} BASE: mono-aug-single & 8.0 & 68.5 \\ 
\rowcolor{LightGray} ETHZ-00-1 & 8.0 & 68.4 \\ 
\rowcolor{LightGray} BASE: trm-aug-shared & 8.0 & 68.0 \\ 
\rowcolor{LightGray} BASE: trm-aug-single & 8.0 & 68.0 \\ 
\rowcolor{LightGray} cmu\_tartan\_01-1 & 8.0 & 68.0 \\ 
\rowcolor{LightGray} NYU-CUBoulder-01-0 & 8.0 & 67.9 \\ 
\rowcolor{LightGray} BASE: trm-shared & 8.0 & 67.7 \\ 
\rowcolor{LightGray} BASE: trm-single & 8.0 & 67.7 \\ 
\rowcolor{LightGray} cmu\_tartan\_00-0 & 8.0 & 67.6 \\ 
\rowcolor{LightGray} cmu\_tartan\_01-0 & 8.0 & 67.6 \\ 
\rowcolor{LightGray} BASE: mono-shared & 8.0 & 66.8 \\ 
\rowcolor{LightGray} BASE: mono-single & 8.0 & 66.8 \\ 
\rowcolor{LightGray} NYU-CUBoulder-02-0 & 8.0 & 66.5 \\ 
\rowcolor{LightGray} deepspin-02-1 & 8.0 & 66.5 \\ 
deepspin-01-1 & 24.0 & 65.1 \\ 
NYU-CUBoulder-03-0 & 24.0 & 64.7 \\ 
\rowcolor{LightGray} NYU-CUBoulder-04-0 & 26.0 & 61.8 \\ 
CU7565-02-0 & 27.0 & 55.5 \\ 
\rowcolor{LightGray} LTI-00-1 & 28.0 & 44.9 \\ 
\rowcolor{LightGray} flexica-01-1 & 28.0 & 41.5 \\ 
\textit{*CU7565-01-0} & \textit{30.0} & \textit{0.0} \\ 
\midrule
Oracle (Baselines) & & 86.9 \\
Oracle (Submissions) & & 98.7 \\
Oracle (All) & & 98.8 \\
\bottomrule
\end{tabular}
%\caption{Results on the Algic genus (1 languages).}
%\label{tab:Algic}
%\end{table}

 \caption{Results on the Algic family (1 language)}
 \end{subtable}
\end{table}

\begin{table}[!htb]
\caption{\textbf{Results per Language Family: Austronesian and Dravidian}}
\centering
 \begin{subtable}{.5\linewidth}
  \centering
%\begin{table}[t]
%\centering
\begin{tabular}{c|c|c}
\toprule
System & Rank & Acc \\ \midrule
CULing-01-0 & 1.0 & 84.4 \\ 
\rowcolor{LightGray} IMS-00-0 & 1.6 & 85.1 \\ 
\rowcolor{LightGray} NYU-CUBoulder-03-0 & 1.6 & 83.6 \\ 
\rowcolor{LightGray} ETHZ-00-1 & 1.6 & 83.4 \\ 
\rowcolor{LightGray} NYU-CUBoulder-01-0 & 1.6 & 82.9 \\ 
\rowcolor{LightGray} NYU-CUBoulder-04-0 & 1.6 & 82.9 \\ 
\rowcolor{LightGray} BASE: trm-shared & 1.6 & 82.8 \\ 
\rowcolor{LightGray} NYU-CUBoulder-02-0 & 1.6 & 82.7 \\ 
deepspin-02-1 & 3.2 & 82.4 \\ 
BASE: trm-aug-single & 3.2 & 81.6 \\ 
\rowcolor{LightGray} \textit{*CU7565-01-0} & \textit{6.8} & \textit{82.7} \\ 
uiuc-01-0 & 5.4 & 82.3 \\ 
\rowcolor{LightGray} BASE: trm-single & 6.0 & 81.2 \\ 
BASE: mono-aug-shared & 6.0 & 82.9 \\ 
LTI-00-1 & 6.0 & 82.0 \\ 
\rowcolor{LightGray} BASE: mono-aug-single & 7.8 & 81.3 \\ 
deepspin-01-1 & 7.6 & 81.0 \\ 
\rowcolor{LightGray} BASE: trm-aug-shared & 7.6 & 79.8 \\ 
\rowcolor{LightGray} flexica-03-1 & 7.6 & 79.3 \\ 
cmu\_tartan\_00-0 & 8.2 & 79.1 \\ 
\rowcolor{LightGray} BASE: mono-shared & 10.4 & 79.2 \\ 
\rowcolor{LightGray} BASE: mono-single & 10.4 & 77.6 \\ 
cmu\_tartan\_00-1 & 12.8 & 80.3 \\ 
cmu\_tartan\_02-1 & 12.8 & 78.9 \\ 
cmu\_tartan\_01-0 & 12.8 & 78.6 \\ 
flexica-02-1 & 12.8 & 78.3 \\ 
cmu\_tartan\_01-1 & 12.8 & 78.2 \\ 
\rowcolor{LightGray} ETHZ-02-1 & 12.0 & 77.4 \\ 
\textit{*CU7565-02-0} & \textit{22.4} & \textit{73.7} \\ 
\rowcolor{LightGray} flexica-01-1 & 21.2 & 69.7 \\ 
\midrule
Oracle (Baselines) & & 89.1 \\
Oracle (Submissions) & & 93.5 \\
Oracle (All) & & 93.7 \\
\bottomrule
\end{tabular}
%%\caption{Results on the Austronesian genus (5 languages).}
%\label{tab:Austronesian}
%\end{table}

 \caption{Results on the Austronesian family (5 languages)}
 \end{subtable}%
 \begin{subtable}{.5\linewidth}
  \centering
%\begin{table}[t]
%\centering
\begin{tabular}{c|c|c}
\toprule
System & Rank & Acc \\ \midrule
IMS-00-0 & 1.0 & 87.6 \\ 
CULing-01-0 & 1.0 & 87.0 \\ 
BASE: trm-aug-shared & 1.0 & 86.8 \\ 
cmu\_tartan\_00-0 & 1.0 & 86.3 \\ 
cmu\_tartan\_01-1 & 1.0 & 86.3 \\ 
BASE: trm-aug-single & 1.0 & 85.9 \\ 
BASE: trm-shared & 1.0 & 85.8 \\ 
ETHZ-02-1 & 1.0 & 85.5 \\ 
\rowcolor{LightGray} cmu\_tartan\_01-0 & 5.0 & 85.7 \\ 
\rowcolor{LightGray} deepspin-02-1 & 5.0 & 85.6 \\ 
\rowcolor{LightGray} cmu\_tartan\_02-1 & 5.0 & 85.5 \\ 
\rowcolor{LightGray} BASE: trm-single & 5.0 & 85.4 \\ 
\rowcolor{LightGray} uiuc-01-0 & 5.0 & 85.3 \\ 
\rowcolor{LightGray} deepspin-01-1 & 5.0 & 85.2 \\ 
\rowcolor{LightGray} LTI-00-1 & 5.0 & 85.0 \\ 
\rowcolor{LightGray} ETHZ-00-1 & 5.0 & 84.9 \\ 
\rowcolor{LightGray} BASE: mono-single & 5.0 & 84.8 \\ 
\rowcolor{LightGray} BASE: mono-aug-single & 5.0 & 84.1 \\ 
NYU-CUBoulder-02-0 & 12.0 & 82.2 \\ 
NYU-CUBoulder-01-0 & 12.0 & 82.2 \\ 
NYU-CUBoulder-03-0 & 12.0 & 82.1 \\ 
NYU-CUBoulder-04-0 & 12.0 & 81.9 \\ 
\rowcolor{LightGray} CU7565-02-0 & 14.5 & 81.4 \\ 
flexica-02-1 & 16.5 & 83.7 \\ 
BASE: mono-shared & 16.5 & 83.7 \\ 
flexica-03-1 & 16.5 & 83.0 \\ 
\rowcolor{LightGray} cmu\_tartan\_00-1 & 19.0 & 62.6 \\ 
BASE: mono-aug-shared & 23.5 & 79.7 \\ 
\rowcolor{LightGray} flexica-01-1 & 28.5 & 56.9 \\ 
\textit{*CU7565-01-0} & \textit{30.0} & \textit{0.0} \\ 
\midrule
Oracle (Baselines) & & 95.9 \\
Oracle (Submissions) & & 98.2 \\
Oracle (All) & & 98.6 \\
\bottomrule
\end{tabular}
%\caption{Results on the Dravidian genus (2 languages).}
%\label{tab:Dravidian}
%\end{table}

\caption{Results on the Dravidian family (2 languages)}
 \end{subtable}
\end{table}

\begin{table}[!htb]
\caption{\textbf{Results per Language Family: Indo-European and Niger-Congo}}
\centering
 \begin{subtable}{.5\linewidth}
  \centering
%\begin{table}[t]
%\centering
\begin{tabular}{c|c|c}
\toprule
System & Rank & Acc \\ \midrule
deepspin-02-1 & 2.3 & 92.9 \\ 
\rowcolor{LightGray} uiuc-01-0 & 3.1 & 91.6 \\ 
deepspin-01-1 & 2.9 & 92.9 \\ 
\rowcolor{LightGray} BASE: trm-single & 2.9 & 91.7 \\ 
CULing-01-0 & 3.9 & 93.5 \\ 
\rowcolor{LightGray} BASE: trm-aug-single & 3.4 & 92.9 \\ 
NYU-CUBoulder-04-0 & 7.3 & 90.7 \\ 
\rowcolor{LightGray} BASE: trm-shared & 12.0 & 86.9 \\ 
cmu\_tartan\_00-1 & 8.1 & 88.6 \\ 
\rowcolor{LightGray} BASE: mono-shared & 8.9 & 90.3 \\ 
NYU-CUBoulder-03-0 & 10.0 & 91.2 \\ 
\rowcolor{LightGray} cmu\_tartan\_00-0 & 8.9 & 88.5 \\ 
NYU-CUBoulder-02-0 & 11.4 & 90.6 \\ 
\rowcolor{LightGray} BASE: mono-aug-shared & 12.9 & 90.5 \\ 
NYU-CUBoulder-01-0 & 12.4 & 90.4 \\ 
\rowcolor{LightGray} BASE: mono-single & 8.1 & 88.0 \\ 
BASE: mono-aug-single & 7.9 & 91.9 \\ 
\rowcolor{LightGray} cmu\_tartan\_01-0 & 10.5 & 88.6 \\ 
cmu\_tartan\_01-1 & 9.9 & 88.5 \\ 
\rowcolor{LightGray} IMS-00-0 & 15.9 & 90.4 \\ 
cmu\_tartan\_02-1 & 10.7 & 88.4 \\ 
\rowcolor{LightGray} BASE: trm-aug-shared & 15.0 & 88.6 \\ 
LTI-00-1 & 15.8 & 87.5 \\ 
\rowcolor{LightGray} CU7565-02-0 & 20.3 & 86.3 \\ 
flexica-03-1 & 19.4 & 80.7 \\ 
\rowcolor{LightGray} ETHZ-02-1 & 18.1 & 83.8 \\ 
ETHZ-00-1 & 23.5 & 73.7 \\ 
\rowcolor{LightGray} flexica-02-1 & 21.8 & 77.5 \\ 
\textit{*CU7565-01-0} & \textit{28.8} & \textit{91.4} \\ 
\rowcolor{LightGray} flexica-01-1 & 26.0 & 76.7 \\ 
\midrule
Oracle (Baselines) & & 98.0 \\
Oracle (Submissions) & & 98.8 \\
Oracle (All) & & 99.1 \\
\bottomrule
\end{tabular}
%\caption{Results on the Indo-European genus (28 languages).}
%\label{tab:IndoEuropean}
%\end{table}

 \caption{Results on the Indo-European family (28 languages)}
 \end{subtable}%
 \begin{subtable}{.5\linewidth}
  \centering
%\begin{table}[t]
%\centering
\begin{tabular}{c|c|c}
\toprule
System & Rank & Acc \\ \midrule
IMS-00-0 & 1.0 & 98.1 \\ 
uiuc-01-0 & 1.0 & 97.9 \\ 
\rowcolor{LightGray} NYU-CUBoulder-01-0 & 1.3 & 98.1 \\ 
\rowcolor{LightGray} NYU-CUBoulder-02-0 & 1.3 & 98.1 \\ 
\rowcolor{LightGray} deepspin-02-1 & 1.3 & 98.0 \\ 
\rowcolor{LightGray} NYU-CUBoulder-03-0 & 1.3 & 98.0 \\ 
\rowcolor{LightGray} BASE: mono-aug-single & 1.3 & 97.9 \\ 
\rowcolor{LightGray} deepspin-01-1 & 1.3 & 97.9 \\ 
\rowcolor{LightGray} NYU-CUBoulder-04-0 & 1.3 & 97.8 \\ 
\rowcolor{LightGray} LTI-00-1 & 1.3 & 97.7 \\ 
\rowcolor{LightGray} BASE: trm-shared & 1.3 & 97.7 \\ 
\rowcolor{LightGray} BASE: trm-single & 1.3 & 97.7 \\ 
\rowcolor{LightGray} BASE: mono-single & 1.3 & 97.7 \\ 
\rowcolor{LightGray} BASE: mono-shared & 1.3 & 97.6 \\ 
\rowcolor{LightGray} BASE: trm-aug-single & 1.3 & 97.5 \\ 
\rowcolor{LightGray} BASE: trm-aug-shared & 1.3 & 97.4 \\ 
\rowcolor{LightGray} BASE: mono-aug-shared & 1.3 & 97.2 \\ 
\textit{*CU7565-01-0} & \textit{3.9} & \textit{98.0} \\ 
\rowcolor{LightGray} CULing-01-0 & 3.4 & 97.1 \\ 
flexica-03-1 & 3.1 & 96.9 \\ 
flexica-02-1 & 3.1 & 96.9 \\ 
\rowcolor{LightGray} cmu\_tartan\_01-1 & 3.6 & 96.4 \\ 
\rowcolor{LightGray} cmu\_tartan\_00-0 & 3.6 & 96.3 \\ 
\rowcolor{LightGray} cmu\_tartan\_01-0 & 3.6 & 96.3 \\ 
CU7565-02-0 & 6.5 & 95.6 \\ 
\rowcolor{LightGray} cmu\_tartan\_00-1 & 7.8 & 95.4 \\ 
flexica-01-1 & 9.2 & 94.2 \\ 
\rowcolor{LightGray} cmu\_tartan\_02-1 & 11.2 & 94.4 \\ 
ETHZ-02-1 & 18.9 & 91.7 \\ 
\rowcolor{LightGray} ETHZ-00-1 & 20.3 & 89.3 \\ 
\midrule
Oracle (Baselines) & & 99.2 \\
Oracle (Submissions) & & 99.4 \\
Oracle (All) & & 99.6 \\
\bottomrule
\end{tabular}
%\caption{Results on the Niger-Congo genus (10 languages).}
%\label{tab:NigerCongo}
%\end{table}

\caption{Results on the Niger-Congo family (10 languages)}
 \end{subtable}
\end{table}

\begin{table}[!htb]
\caption{\textbf{Results per Language Family: Oto-Manguean and Sino-Tibetan}}
\centering
 \begin{subtable}{.5\linewidth}
  \centering
%\begin{table}[t]
%\centering
\begin{tabular}{c|c|c}
\toprule
System & Rank & Acc \\ \midrule
uiuc-01-0 & 1.0 & 87.5 \\ 
\rowcolor{LightGray} BASE: trm-single & 2.0 & 86.2 \\ 
CULing-01-0 & 3.1 & 86.7 \\ 
\rowcolor{LightGray} deepspin-02-1 & 3.4 & 85.4 \\ 
\rowcolor{LightGray} deepspin-01-1 & 3.4 & 85.3 \\ 
NYU-CUBoulder-04-0 & 6.4 & 84.2 \\ 
\rowcolor{LightGray} BASE: mono-single & 7.9 & 82.4 \\ 
NYU-CUBoulder-03-0 & 8.4 & 83.5 \\ 
\rowcolor{LightGray} BASE: mono-aug-single & 6.1 & 83.5 \\ 
BASE: mono-shared & 8.2 & 82.9 \\ 
\rowcolor{LightGray} NYU-CUBoulder-02-0 & 9.1 & 83.5 \\ 
IMS-00-0 & 10.3 & 83.3 \\ 
\rowcolor{LightGray} LTI-00-1 & 9.4 & 82.4 \\ 
NYU-CUBoulder-01-0 & 9.4 & 83.6 \\ 
\rowcolor{LightGray} BASE: mono-aug-shared & 9.8 & 82.0 \\ 
cmu\_tartan\_00-0 & 13.9 & 78.5 \\ 
\rowcolor{LightGray} cmu\_tartan\_01-1 & 14.9 & 78.5 \\ 
cmu\_tartan\_02-1 & 15.2 & 78.2 \\ 
\rowcolor{LightGray} BASE: trm-shared & 14.5 & 80.2 \\ 
BASE: trm-aug-shared & 20.3 & 73.8 \\ 
\rowcolor{LightGray} flexica-01-1 & 26.3 & 47.2 \\ 
BASE: trm-aug-single & 7.4 & 84.3 \\ 
\rowcolor{LightGray} cmu\_tartan\_00-1 & 14.1 & 79.0 \\ 
ETHZ-02-1 & 14.0 & 81.4 \\ 
\rowcolor{LightGray} CU7565-02-0 & 20.9 & 75.1 \\ 
cmu\_tartan\_01-0 & 18.3 & 76.5 \\ 
\rowcolor{LightGray} \textit{*CU7565-01-0} & \textit{27.8} & \textit{81.0} \\ 
ETHZ-00-1 & 25.4 & 70.5 \\ 
\rowcolor{LightGray} flexica-02-1 & 25.6 & 67.0 \\ 
flexica-03-1 & 26.1 & 64.2 \\ 
\midrule
Oracle (Baselines) & & 94.1 \\
Oracle (Submissions) & & 96.2 \\
Oracle (All) & & 96.7 \\
\bottomrule
\end{tabular}
%\caption{Results on the Oto-Manguean genus (10 languages).}
%\label{tab:OtoManguean}
%\end{table}

 \caption{Results on the Oto-Manguean family (10 languages)}
 \end{subtable}%
 \begin{subtable}{.5\linewidth}
  \centering
%\begin{table}[t]
%\centering
\begin{tabular}{c|c|c}
\toprule
System & Rank & Acc \\ \midrule
deepspin-01-1 & 1.0 & 85.1 \\ 
deepspin-02-1 & 1.0 & 85.0 \\ 
LTI-00-1 & 1.0 & 84.7 \\ 
uiuc-01-0 & 1.0 & 84.4 \\ 
BASE: trm-single & 1.0 & 84.4 \\ 
BASE: trm-shared & 1.0 & 84.4 \\ 
CULing-01-0 & 1.0 & 84.1 \\ 
ETHZ-02-1 & 1.0 & 83.8 \\ 
flexica-02-1 & 1.0 & 83.7 \\ 
cmu\_tartan\_01-1 & 1.0 & 83.4 \\ 
BASE: mono-aug-shared & 1.0 & 83.4 \\ 
BASE: mono-aug-single & 1.0 & 83.4 \\ 
NYU-CUBoulder-01-0 & 1.0 & 83.4 \\ 
IMS-00-0 & 1.0 & 83.3 \\ 
BASE: trm-aug-single & 1.0 & 83.3 \\ 
BASE: trm-aug-shared & 1.0 & 83.3 \\ 
BASE: mono-shared & 1.0 & 83.2 \\ 
BASE: mono-single & 1.0 & 83.2 \\ 
cmu\_tartan\_00-0 & 1.0 & 83.1 \\ 
cmu\_tartan\_02-1 & 1.0 & 83.1 \\ 
cmu\_tartan\_00-1 & 1.0 & 83.0 \\ 
\rowcolor{LightGray} NYU-CUBoulder-03-0 & 22.0 & 82.8 \\ 
\rowcolor{LightGray} ETHZ-00-1 & 22.0 & 82.8 \\ 
\rowcolor{LightGray} cmu\_tartan\_01-0 & 22.0 & 82.7 \\ 
\rowcolor{LightGray} NYU-CUBoulder-02-0 & 22.0 & 82.6 \\ 
\rowcolor{LightGray} flexica-03-1 & 22.0 & 82.5 \\ 
\rowcolor{LightGray} NYU-CUBoulder-04-0 & 22.0 & 81.7 \\ 
flexica-01-1 & 28.0 & 70.6 \\ 
CU7565-02-0 & 28.0 & 67.9 \\ 
\rowcolor{LightGray} \textit{*CU7565-01-0} & \textit{30.0} & \textit{0.0} \\ 
\midrule
Oracle (Baselines) & & 91.3 \\
Oracle (Submissions) & & 96.0 \\
Oracle (All) & & 96.2 \\
\bottomrule
\end{tabular}
%\caption{Results on the Sino-Tibetan genus (1 languages).}
%\label{tab:SinoTibetan}
%\end{table}

\caption{Results on the Sino-Tibetan family (1 language)}
 \end{subtable}
\end{table}

\begin{table}[!htb]
\caption{\textbf{Results per Language Family: Siouan and Songhay}}
\centering
 \begin{subtable}{.5\linewidth}
  \centering
%\begin{table}[t]
%\centering
\begin{tabular}{c|c|c}
\toprule
System & Rank & Acc \\ \midrule
NYU-CUBoulder-01-0 & 1.0 & 95.7 \\ 
BASE: trm-single & 1.0 & 95.6 \\ 
CULing-01-0 & 1.0 & 95.6 \\ 
BASE: trm-shared & 1.0 & 95.6 \\ 
ETHZ-00-1 & 1.0 & 95.5 \\ 
uiuc-01-0 & 1.0 & 94.9 \\ 
deepspin-01-1 & 1.0 & 94.8 \\ 
NYU-CUBoulder-02-0 & 1.0 & 94.8 \\ 
NYU-CUBoulder-03-0 & 1.0 & 94.7 \\ 
deepspin-02-1 & 1.0 & 94.5 \\ 
BASE: mono-aug-shared & 1.0 & 94.4 \\ 
BASE: mono-aug-single & 1.0 & 94.4 \\ 
NYU-CUBoulder-04-0 & 1.0 & 94.3 \\ 
\rowcolor{LightGray} ETHZ-02-1 & 14.0 & 93.3 \\ 
\rowcolor{LightGray} BASE: mono-single & 14.0 & 92.9 \\ 
\rowcolor{LightGray} BASE: mono-shared & 14.0 & 92.9 \\ 
\rowcolor{LightGray} BASE: trm-aug-single & 14.0 & 92.5 \\ 
\rowcolor{LightGray} BASE: trm-aug-shared & 14.0 & 92.5 \\ 
\rowcolor{LightGray} flexica-02-1 & 14.0 & 91.5 \\ 
\rowcolor{LightGray} IMS-00-0 & 14.0 & 90.9 \\ 
LTI-00-1 & 21.0 & 89.7 \\ 
flexica-03-1 & 21.0 & 89.3 \\ 
\rowcolor{LightGray} cmu\_tartan\_01-0 & 23.0 & 85.7 \\ 
\rowcolor{LightGray} cmu\_tartan\_01-1 & 23.0 & 85.7 \\ 
\rowcolor{LightGray} cmu\_tartan\_02-1 & 23.0 & 85.7 \\ 
\rowcolor{LightGray} cmu\_tartan\_00-0 & 23.0 & 85.5 \\ 
\rowcolor{LightGray} cmu\_tartan\_00-1 & 23.0 & 85.5 \\ 
CU7565-02-0 & 28.0 & 80.5 \\ 
\rowcolor{LightGray} flexica-01-1 & 29.0 & 58.4 \\ 
\textit{*CU7565-01-0} & \textit{30.0} & \textit{0.0} \\ 
\midrule
Oracle (Baselines) & & 97.3 \\
Oracle (Submissions) & & 98.1 \\
Oracle (All) & & 98.1 \\
\bottomrule
\end{tabular}
%\caption{Results on the Siouan genus (1 languages).}
%\label{tab:Siouan}
%\end{table}

 \caption{Results on the Siouan family (1 language)}
 \end{subtable}%
 \begin{subtable}{.5\linewidth}
  \centering
%\begin{table}[t]
%\centering
\begin{tabular}{c|c|c}
\toprule
System & Rank & Acc \\ \midrule
BASE: mono-aug-single & 1.0 & 100.0 \\ 
BASE: trm-aug-single & 1.0 & 100.0 \\ 
CU7565-02-0 & 1.0 & 100.0 \\ 
CU7565-01-0 & 1.0 & 100.0 \\ 
uiuc-01-0 & 1.0 & 100.0 \\ 
NYU-CUBoulder-02-0 & 1.0 & 100.0 \\ 
NYU-CUBoulder-03-0 & 1.0 & 100.0 \\ 
BASE: mono-aug-shared & 1.0 & 100.0 \\ 
NYU-CUBoulder-01-0 & 1.0 & 100.0 \\ 
LTI-00-1 & 1.0 & 100.0 \\ 
IMS-00-0 & 1.0 & 100.0 \\ 
flexica-01-1 & 1.0 & 100.0 \\ 
deepspin-02-1 & 1.0 & 100.0 \\ 
deepspin-01-1 & 1.0 & 100.0 \\ 
CULing-01-0 & 1.0 & 100.0 \\ 
cmu\_tartan\_01-1 & 1.0 & 100.0 \\ 
NYU-CUBoulder-04-0 & 1.0 & 100.0 \\ 
BASE: trm-aug-shared & 1.0 & 100.0 \\ 
flexica-03-1 & 1.0 & 93.8 \\ 
ETHZ-00-1 & 1.0 & 93.8 \\ 
cmu\_tartan\_02-1 & 1.0 & 93.8 \\ 
cmu\_tartan\_01-0 & 1.0 & 93.8 \\ 
cmu\_tartan\_00-0 & 1.0 & 87.5 \\ 
cmu\_tartan\_00-1 & 1.0 & 87.5 \\ 
BASE: trm-shared & 1.0 & 87.5 \\ 
BASE: trm-single & 1.0 & 87.5 \\ 
\rowcolor{LightGray} flexica-02-1 & 27.0 & 0.0 \\ 
\rowcolor{LightGray} BASE: mono-shared & 27.0 & 0.0 \\ 
\rowcolor{LightGray} BASE: mono-single & 27.0 & 0.0 \\ 
\rowcolor{LightGray} ETHZ-02-1 & 27.0 & 0.0 \\ 
\midrule
Oracle (Baselines) & & 100.0 \\
Oracle (Submissions) & & 100.0 \\
Oracle (All) & & 100.0 \\
\bottomrule
\end{tabular}
%\caption{Results on the Songhay genus (1 languages).}
%\label{tab:Songhay}
%\end{table}

\caption{Results on the Songhay family/genus (1 language)}
 \end{subtable}
\end{table}

\begin{table}[!htb]
\caption{\textbf{Results per Language Family: Southern Daly and Tungusic}}
\centering
 \begin{subtable}{.5\linewidth}
  \centering
%\begin{table}[t]
%\centering
\begin{tabular}{c|c|c}
\toprule
System & Rank & Acc \\ \midrule
CULing-01-0 & 1.0 & 91.9 \\ 
BASE: trm-single & 1.0 & 89.6 \\ 
BASE: trm-shared & 1.0 & 89.6 \\ 
ETHZ-00-1 & 1.0 & 88.7 \\ 
uiuc-01-0 & 1.0 & 87.8 \\ 
BASE: trm-aug-single & 1.0 & 86.9 \\ 
BASE: trm-aug-shared & 1.0 & 86.9 \\ 
IMS-00-0 & 1.0 & 86.0 \\ 
\rowcolor{LightGray} deepspin-01-1 & 9.0 & 83.8 \\ 
\rowcolor{LightGray} deepspin-02-1 & 9.0 & 83.3 \\ 
\rowcolor{LightGray} cmu\_tartan\_01-1 & 9.0 & 81.1 \\ 
\rowcolor{LightGray} cmu\_tartan\_01-0 & 9.0 & 81.1 \\ 
\rowcolor{LightGray} cmu\_tartan\_00-0 & 9.0 & 80.2 \\ 
\rowcolor{LightGray} cmu\_tartan\_00-1 & 9.0 & 80.2 \\ 
ETHZ-02-1 & 15.0 & 77.9 \\ 
CU7565-02-0 & 15.0 & 77.5 \\ 
flexica-03-1 & 15.0 & 73.4 \\ 
flexica-02-1 & 15.0 & 72.5 \\ 
LTI-00-1 & 15.0 & 70.3 \\ 
\rowcolor{LightGray} cmu\_tartan\_02-1 & 20.0 & 67.1 \\ 
\rowcolor{LightGray} BASE: mono-shared & 20.0 & 60.8 \\ 
\rowcolor{LightGray} BASE: mono-single & 20.0 & 60.8 \\ 
\rowcolor{LightGray} NYU-CUBoulder-04-0 & 20.0 & 59.5 \\ 
\rowcolor{LightGray} NYU-CUBoulder-03-0 & 20.0 & 59.0 \\ 
\rowcolor{LightGray} NYU-CUBoulder-02-0 & 20.0 & 57.7 \\ 
\rowcolor{LightGray} NYU-CUBoulder-01-0 & 20.0 & 57.7 \\ 
BASE: mono-aug-single & 27.0 & 44.6 \\ 
BASE: mono-aug-shared & 27.0 & 44.6 \\ 
\rowcolor{LightGray} flexica-01-1 & 29.0 & 21.2 \\ 
\textit{*CU7565-01-0} & \textit{30.0} & \textit{0.0} \\ 
\midrule
Oracle (Baselines) & & 91.4 \\
Oracle (Submissions) & & 96.4 \\
Oracle (All) & & 96.4 \\
\bottomrule
\end{tabular}
%\caption{Results on the Southern Daly genus (1 languages).}
%\label{tab:SouthernDaly}
%\end{table}

 \caption{Results on the Southern Daly family (1 language)}
 \end{subtable}%
 \begin{subtable}{.5\linewidth}
  \centering
%\begin{table}[t]
%\centering
\begin{tabular}{c|c|c}
\toprule
System & Rank & Acc \\ \midrule
deepspin-02-1 & 1.0 & 59.0 \\ 
deepspin-01-1 & 1.0 & 58.8 \\ 
uiuc-01-0 & 1.0 & 58.3 \\ 
IMS-00-0 & 1.0 & 58.2 \\ 
CULing-01-0 & 1.0 & 58.0 \\ 
BASE: trm-aug-single & 1.0 & 57.7 \\ 
BASE: trm-aug-shared & 1.0 & 57.7 \\ 
ETHZ-00-1 & 1.0 & 57.2 \\ 
BASE: trm-single & 1.0 & 57.1 \\ 
cmu\_tartan\_01-0 & 1.0 & 57.1 \\ 
BASE: trm-shared & 1.0 & 57.1 \\ 
\rowcolor{LightGray} cmu\_tartan\_00-0 & 12.0 & 56.8 \\ 
\rowcolor{LightGray} cmu\_tartan\_01-1 & 12.0 & 56.5 \\ 
\rowcolor{LightGray} cmu\_tartan\_00-1 & 12.0 & 55.9 \\ 
\rowcolor{LightGray} LTI-00-1 & 12.0 & 55.0 \\ 
cmu\_tartan\_02-1 & 16.0 & 54.1 \\ 
BASE: mono-single & 16.0 & 54.0 \\ 
BASE: mono-shared & 16.0 & 54.0 \\ 
ETHZ-02-1 & 16.0 & 53.6 \\ 
BASE: mono-aug-single & 16.0 & 53.5 \\ 
BASE: mono-aug-shared & 16.0 & 53.5 \\ 
flexica-02-1 & 16.0 & 53.1 \\ 
flexica-03-1 & 16.0 & 52.7 \\ 
\rowcolor{LightGray} NYU-CUBoulder-01-0 & 24.0 & 50.0 \\ 
\rowcolor{LightGray} NYU-CUBoulder-03-0 & 24.0 & 48.8 \\ 
\rowcolor{LightGray} NYU-CUBoulder-02-0 & 24.0 & 48.6 \\ 
\rowcolor{LightGray} NYU-CUBoulder-04-0 & 24.0 & 48.2 \\ 
flexica-01-1 & 28.0 & 46.5 \\ 
\rowcolor{LightGray} CU7565-02-0 & 29.0 & 43.5 \\ 
\textit{*CU7565-01-0} & \textit{30.0} & \textit{0.0} \\ 
\midrule
Oracle (Baselines) & & 67.7 \\
Oracle (Submissions) & & 75.9 \\
Oracle (All) & & 76.3 \\
\bottomrule
\end{tabular}
%\caption{Results on the Tungusic genus (1 languages).}
%\label{tab:Tungusic}
%\end{table}

\caption{Results on the Tungusic family (1 language)}
 \end{subtable}
\end{table}

\begin{table}[!htb]
\caption{\textbf{Results per Language Family: Turkic and Uralic}}
\centering
 \begin{subtable}{.5\linewidth}
  \centering
%\begin{table}[t]
%\centering
\begin{tabular}{c|c|c}
\toprule
System & Rank & Acc \\ \midrule
BASE: trm-single & 1.0 & 91.8 \\ 
BASE: trm-aug-single & 1.0 & 91.8 \\ 
\rowcolor{LightGray} uiuc-01-0 & 1.8 & 92.0 \\ 
CULing-01-0 & 3.5 & 91.9 \\ 
\rowcolor{LightGray} deepspin-02-1 & 6.7 & 91.3 \\ 
\rowcolor{LightGray} deepspin-01-1 & 6.7 & 91.1 \\ 
NYU-CUBoulder-04-0 & 5.5 & 90.4 \\ 
\rowcolor{LightGray} BASE: mono-single & 5.1 & 90.9 \\ 
NYU-CUBoulder-02-0 & 6.8 & 90.6 \\ 
NYU-CUBoulder-03-0 & 6.8 & 90.5 \\ 
\rowcolor{LightGray} cmu\_tartan\_01-1 & 7.2 & 91.0 \\ 
cmu\_tartan\_00-1 & 6.6 & 90.8 \\ 
\rowcolor{LightGray} BASE: mono-aug-single & 7.3 & 90.7 \\ 
BASE: trm-shared & 7.7 & 91.3 \\ 
\rowcolor{LightGray} cmu\_tartan\_02-1 & 7.4 & 90.8 \\ 
NYU-CUBoulder-01-0 & 8.9 & 90.5 \\ 
\rowcolor{LightGray} BASE: trm-aug-shared & 9.3 & 91.1 \\ 
cmu\_tartan\_00-0 & 9.7 & 90.9 \\ 
\rowcolor{LightGray} cmu\_tartan\_01-0 & 11.8 & 90.7 \\ 
ETHZ-00-1 & 16.6 & 88.9 \\ 
\rowcolor{LightGray} IMS-00-0 & 11.2 & 91.0 \\ 
BASE: mono-shared & 15.1 & 88.9 \\ 
\rowcolor{LightGray} flexica-02-1 & 13.1 & 89.7 \\ 
LTI-00-1 & 17.1 & 83.3 \\ 
\rowcolor{LightGray} flexica-03-1 & 17.0 & 88.6 \\ 
BASE: mono-aug-shared & 19.5 & 86.3 \\ 
\rowcolor{LightGray} CU7565-02-0 & 21.6 & 85.9 \\ 
ETHZ-02-1 & 17.5 & 88.6 \\ 
\rowcolor{LightGray} \textit{*CU7565-01-0} & \textit{29.1} & \textit{96.4} \\ 
flexica-01-1 & 28.9 & 72.4 \\ 
\midrule
Oracle (Baselines) & & 95.8 \\
Oracle (Submissions) & & 97.4 \\
Oracle (All) & & 97.5 \\
\bottomrule
\end{tabular}
%\caption{Results on the Turkic genus (10 languages).}
%\label{tab:Turkic}
%\end{table}

 \caption{Results on the Turkic family (10 languages)}
 \end{subtable}%
 \begin{subtable}{.5\linewidth}
  \centering
%\begin{table}[t]
%\centering
\begin{tabular}{c|c|c}
\toprule
System & Rank & Acc \\ \midrule
deepspin-02-1 & 1.8 & 90.7 \\ 
\rowcolor{LightGray} deepspin-01-1 & 3.1 & 89.7 \\ 
uiuc-01-0 & 2.8 & 88.2 \\ 
\rowcolor{LightGray} CULing-01-0 & 3.9 & 88.9 \\ 
BASE: trm-single & 3.8 & 88.1 \\ 
\rowcolor{LightGray} BASE: trm-aug-single & 4.3 & 88.5 \\ 
NYU-CUBoulder-04-0 & 10.6 & 86.8 \\ 
\rowcolor{LightGray} NYU-CUBoulder-02-0 & 13.4 & 86.4 \\ 
NYU-CUBoulder-03-0 & 13.4 & 86.0 \\ 
\rowcolor{LightGray} IMS-00-0 & 14.8 & 86.1 \\ 
NYU-CUBoulder-01-0 & 15.4 & 85.9 \\ 
\rowcolor{LightGray} cmu\_tartan\_00-1 & 7.7 & 85.8 \\ 
cmu\_tartan\_02-1 & 9.8 & 84.8 \\ 
\rowcolor{LightGray} LTI-00-1 & 12.3 & 86.7 \\ 
cmu\_tartan\_01-1 & 7.6 & 86.0 \\ 
\rowcolor{LightGray} cmu\_tartan\_00-0 & 8.7 & 86.2 \\ 
BASE: trm-aug-shared & 18.8 & 82.6 \\ 
\rowcolor{LightGray} \textit{*CU7565-02-0} & \textit{22.2} & \textit{79.4} \\ 
\textit{*CU7565-01-0} & \textit{28.2} & \textit{92.9} \\ 
\rowcolor{LightGray} BASE: mono-single & 10.8 & 83.0 \\ 
cmu\_tartan\_01-0 & 10.6 & 84.8 \\ 
\rowcolor{LightGray} BASE: mono-shared & 17.6 & 81.1 \\ 
BASE: mono-aug-shared & 19.4 & 81.9 \\ 
\rowcolor{LightGray} BASE: trm-shared & 19.5 & 76.8 \\ 
ETHZ-02-1 & 22.6 & 67.9 \\ 
\rowcolor{LightGray} BASE: mono-aug-single & 11.4 & 85.9 \\ 
flexica-02-1 & 19.5 & 70.7 \\ 
\rowcolor{LightGray} flexica-03-1 & 20.5 & 67.8 \\ 
flexica-01-1 & 26.8 & 66.0 \\ 
\rowcolor{LightGray} ETHZ-00-1 & 28.3 & 54.9 \\ 
\midrule
Oracle (Baselines) & & 95.5 \\
Oracle (Submissions) & & 96.8 \\
Oracle (All) & & 97.2 \\
\bottomrule
\end{tabular}
%\caption{Results on the Uralic genus (16 languages).}
%\label{tab:Uralic}
%\end{table}

\caption{Results on the Uralic family (16 languages)}
 \end{subtable}
\end{table}

\begin{table}[!htb]
\caption{\textbf{Results per Language Family (Uto-Aztecan) and Semitic Genus (Afro-Asiatic Family)}}
\centering
 \begin{subtable}{.5\linewidth}
  \centering
%\begin{table}[t]
%\centering
\begin{tabular}{c|c|c}
\toprule
System & Rank & Acc \\ \midrule
uiuc-01-0 & 1.0 & 82.5 \\ 
NYU-CUBoulder-01-0 & 1.0 & 82.2 \\ 
NYU-CUBoulder-02-0 & 1.0 & 81.8 \\ 
NYU-CUBoulder-03-0 & 1.0 & 81.5 \\ 
IMS-00-0 & 1.0 & 81.5 \\ 
BASE: trm-single & 1.0 & 80.9 \\ 
CULing-01-0 & 1.0 & 80.9 \\ 
BASE: trm-shared & 1.0 & 80.9 \\ 
deepspin-02-1 & 1.0 & 80.6 \\ 
NYU-CUBoulder-04-0 & 1.0 & 79.6 \\ 
ETHZ-00-1 & 1.0 & 79.3 \\ 
LTI-00-1 & 1.0 & 79.0 \\ 
deepspin-01-1 & 1.0 & 79.0 \\ 
\rowcolor{LightGray} BASE: trm-aug-single & 14.0 & 78.0 \\ 
\rowcolor{LightGray} BASE: trm-aug-shared & 14.0 & 78.0 \\ 
\rowcolor{LightGray} flexica-02-1 & 14.0 & 77.7 \\ 
\rowcolor{LightGray} BASE: mono-aug-single & 14.0 & 77.4 \\ 
\rowcolor{LightGray} BASE: mono-aug-shared & 14.0 & 77.4 \\ 
\rowcolor{LightGray} cmu\_tartan\_00-0 & 14.0 & 76.1 \\ 
\rowcolor{LightGray} cmu\_tartan\_00-1 & 14.0 & 76.1 \\ 
\rowcolor{LightGray} cmu\_tartan\_01-0 & 14.0 & 75.8 \\ 
\rowcolor{LightGray} cmu\_tartan\_01-1 & 14.0 & 75.8 \\ 
\rowcolor{LightGray} BASE: mono-shared & 14.0 & 75.8 \\ 
\rowcolor{LightGray} BASE: mono-single & 14.0 & 75.8 \\ 
\rowcolor{LightGray} flexica-03-1 & 14.0 & 75.5 \\ 
\rowcolor{LightGray} ETHZ-02-1 & 14.0 & 74.5 \\ 
\rowcolor{LightGray} cmu\_tartan\_02-1 & 14.0 & 74.2 \\ 
CU7565-01-0 & 28.0 & 71.0 \\ 
\rowcolor{LightGray} CU7565-02-0 & 29.0 & 62.4 \\ 
flexica-01-1 & 30.0 & 54.8 \\ 
\midrule
Oracle (Baselines) & & 87.2 \\
Oracle (Submissions) & & 92.0 \\
Oracle (All) & & 92.3 \\
\bottomrule
\end{tabular}
%\caption{Results on the Uto-Aztecan genus (1 languages).}
%\label{tab:UtoAztecan}
%\end{table}

 \caption{Results on the Uto-Aztecan family (1 language)}
 \end{subtable}%
 \begin{subtable}{.5\linewidth}
  \centering
 %\begin{table}[t]
%\centering
\begin{tabular}{c|c|c}
\toprule
System & Rank & Acc \\ \midrule
uiuc-01-0 & 1.0 & 95.6 \\ 
CULing-01-0 & 1.0 & 94.9 \\ 
\rowcolor{LightGray} deepspin-02-1 & 5.0 & 93.3 \\ 
BASE: trm-single & 5.5 & 93.9 \\ 
BASE: trm-aug-single & 5.5 & 93.1 \\ 
deepspin-01-1 & 5.5 & 92.5 \\ 
NYU-CUBoulder-01-0 & 5.5 & 92.4 \\ 
NYU-CUBoulder-02-0 & 5.5 & 92.3 \\ 
\rowcolor{LightGray} NYU-CUBoulder-04-0 & 14.0 & 92.0 \\ 
BASE: mono-aug-shared & 9.0 & 91.3 \\ 
BASE: mono-single & 9.0 & 90.2 \\ 
cmu\_tartan\_00-0 & 9.0 & 90.0 \\ 
\rowcolor{LightGray} cmu\_tartan\_01-1 & 13.5 & 85.4 \\ 
\rowcolor{LightGray} cmu\_tartan\_01-0 & 13.5 & 85.2 \\ 
cmu\_tartan\_02-1 & 14.5 & 72.3 \\ 
\rowcolor{LightGray} ETHZ-00-1 & 9.5 & 92.5 \\ 
\rowcolor{LightGray} BASE: trm-aug-shared & 9.5 & 91.8 \\ 
\rowcolor{LightGray} BASE: trm-shared & 9.5 & 91.7 \\ 
\rowcolor{LightGray} IMS-00-0 & 9.5 & 91.7 \\ 
\rowcolor{LightGray} BASE: mono-aug-single & 9.5 & 90.9 \\ 
\rowcolor{LightGray} CU7565-02-0 & 9.5 & 90.6 \\ 
NYU-CUBoulder-03-0 & 18.0 & 91.2 \\ 
\rowcolor{LightGray} LTI-00-1 & 13.5 & 90.1 \\ 
\rowcolor{LightGray} flexica-02-1 & 13.5 & 90.1 \\ 
\rowcolor{LightGray} ETHZ-02-1 & 13.5 & 89.5 \\ 
\rowcolor{LightGray} flexica-03-1 & 13.5 & 89.2 \\ 
\rowcolor{LightGray} cmu\_tartan\_00-1 & 13.5 & 89.0 \\ 
\rowcolor{LightGray} BASE: mono-shared & 13.5 & 88.5 \\ 
\textit{*CU7565-01-0} & \textit{28.5} & \textit{88.3} \\ 
\rowcolor{LightGray} flexica-01-1 & 28.0 & 63.9 \\ 
\midrule
Oracle (Baselines) & & 98.4 \\
Oracle (Submissions) & & 99.6 \\
Oracle (All) & & 99.7 \\
\bottomrule
\end{tabular}
%\caption{Results on the Semitic genus (2 languages).}
%\label{tab:Semitic}
%\end{table}

  \caption{Results on the Semitic genus (2 languages)}
 \end{subtable}
\end{table}

\begin{table}[!htb]
\caption{\textbf{Results per Language Genus (in Indo-European family)}}
\centering
 \begin{subtable}{.5\linewidth}
  \centering
%\begin{table}[t]
%\centering
\begin{tabular}{c|c|c}
\toprule
System & Rank & Acc \\ \midrule
deepspin-02-1 & 3.4 & 87.1 \\ 
\rowcolor{LightGray} deepspin-01-1 & 4.6 & 87.0 \\ 
uiuc-01-0 & 3.5 & 87.4 \\ 
\rowcolor{LightGray} BASE: trm-single & 3.1 & 87.5 \\ 
CULing-01-0 & 3.5 & 88.3 \\ 
\rowcolor{LightGray} BASE: trm-aug-single & 4.9 & 87.4 \\ 
IMS-00-0 & 15.1 & 83.1 \\ 
\rowcolor{LightGray} BASE: mono-single & 5.3 & 86.3 \\ 
BASE: mono-aug-single & 6.8 & 86.3 \\ 
\rowcolor{LightGray} NYU-CUBoulder-04-0 & 10.2 & 85.2 \\ 
NYU-CUBoulder-02-0 & 13.1 & 83.3 \\ 
\rowcolor{LightGray} NYU-CUBoulder-03-0 & 12.0 & 84.4 \\ 
LTI-00-1 & 11.1 & 84.3 \\ 
\rowcolor{LightGray} cmu\_tartan\_00-1 & 9.8 & 79.5 \\ 
NYU-CUBoulder-01-0 & 14.5 & 83.0 \\ 
\rowcolor{LightGray} BASE: mono-aug-shared & 13.2 & 84.4 \\ 
cmu\_tartan\_01-0 & 11.1 & 78.9 \\ 
cmu\_tartan\_01-1 & 11.1 & 78.8 \\ 
\rowcolor{LightGray} cmu\_tartan\_00-0 & 10.8 & 79.3 \\ 
BASE: trm-shared & 19.5 & 77.7 \\ 
\rowcolor{LightGray} BASE: trm-aug-shared & 19.5 & 79.1 \\ 
BASE: mono-shared & 11.7 & 83.7 \\ 
\rowcolor{LightGray} cmu\_tartan\_02-1 & 13.2 & 78.5 \\ 
CU7565-02-0 & 19.4 & 78.6 \\ 
\rowcolor{LightGray} ETHZ-02-1 & 18.9 & 76.4 \\ 
flexica-01-1 & 26.2 & 66.6 \\ 
\rowcolor{LightGray} flexica-03-1 & 25.5 & 66.5 \\ 
flexica-02-1 & 25.9 & 64.2 \\ 
\rowcolor{LightGray} ETHZ-00-1 & 27.1 & 60.1 \\ 
\textit{*CU7565-01-0} & \textit{30.0} & \textit{0.0} \\ 
\midrule
Oracle (Baselines) & & 97.0 \\
Oracle (Submissions) & & 98.4 \\
Oracle (All) & & 98.9 \\
\bottomrule
\end{tabular}
%\caption{Results on the Germanic genus (13 languages).}
%\label{tab:Germanic}
%\end{table}

 \caption{Results on the Germanic genus (13 languages)}
 \end{subtable}%
 \begin{subtable}{.5\linewidth}
  \centering
%\begin{table}[t]
%\centering
\begin{tabular}{c|c|c}
\toprule
System & Rank & Acc \\ \midrule
uiuc-01-0 & 1.0 & 98.2 \\ 
\rowcolor{LightGray} deepspin-02-1 & 1.5 & 98.1 \\ 
\rowcolor{LightGray} deepspin-01-1 & 1.5 & 98.0 \\ 
\rowcolor{LightGray} BASE: trm-single & 1.5 & 97.9 \\ 
\rowcolor{LightGray} BASE: trm-aug-single & 1.5 & 97.8 \\ 
BASE: trm-shared & 2.8 & 97.9 \\ 
\rowcolor{LightGray} CULing-01-0 & 7.5 & 98.0 \\ 
BASE: mono-single & 6.0 & 97.6 \\ 
\rowcolor{LightGray} NYU-CUBoulder-04-0 & 5.0 & 97.7 \\ 
cmu\_tartan\_02-1 & 7.8 & 97.4 \\ 
\rowcolor{LightGray} cmu\_tartan\_00-1 & 7.0 & 97.4 \\ 
\rowcolor{LightGray} BASE: mono-shared & 7.0 & 97.3 \\ 
cmu\_tartan\_01-1 & 7.8 & 97.3 \\ 
\rowcolor{LightGray} cmu\_tartan\_00-0 & 8.8 & 97.1 \\ 
NYU-CUBoulder-03-0 & 8.5 & 97.4 \\ 
\rowcolor{LightGray} NYU-CUBoulder-02-0 & 9.2 & 97.4 \\ 
\rowcolor{LightGray} NYU-CUBoulder-01-0 & 9.2 & 97.3 \\ 
BASE: trm-aug-shared & 11.0 & 97.7 \\ 
\rowcolor{LightGray} BASE: mono-aug-single & 9.5 & 97.2 \\ 
\rowcolor{LightGray} flexica-03-1 & 9.5 & 97.1 \\ 
flexica-02-1 & 11.0 & 96.8 \\ 
\rowcolor{LightGray} ETHZ-02-1 & 11.5 & 97.4 \\ 
ETHZ-00-1 & 13.8 & 96.4 \\ 
\rowcolor{LightGray} BASE: mono-aug-shared & 15.8 & 94.2 \\ 
cmu\_tartan\_01-0 & 17.2 & 96.9 \\ 
\rowcolor{LightGray} IMS-00-0 & 17.0 & 96.6 \\ 
CU7565-02-0 & 19.8 & 94.8 \\ 
\rowcolor{LightGray} LTI-00-1 & 19.8 & 81.5 \\ 
\textit{*CU7565-01-0} & \textit{29.0} & \textit{89.0} \\ 
\rowcolor{LightGray} flexica-01-1 & 28.8 & 88.1 \\ 
\midrule
Oracle (Baselines) & & 99.2 \\
Oracle (Submissions) & & 99.6 \\
Oracle (All) & & 99.7 \\
\bottomrule
\end{tabular}
%\caption{Results on the Indic genus (4 languages).}
%\label{tab:Indic}
%\end{table}

  \caption{Results on the Indic genus (4 languages)}
 \end{subtable}
\end{table}

\begin{table}[!htb]
\caption{\textbf{Results per Language Genus (in Indo-European family)}}
\centering
 \begin{subtable}{.5\linewidth}
  \centering
%\begin{table}[t]
%\centering
\begin{tabular}{c|c|c}
\toprule
System & Rank & Acc \\ \midrule
CULing-01-0 & 1.0 & 95.3 \\ 
\rowcolor{LightGray} deepspin-01-1 & 2.0 & 94.6 \\ 
\rowcolor{LightGray} deepspin-02-1 & 2.0 & 94.6 \\ 
\rowcolor{LightGray} BASE: trm-aug-shared & 2.0 & 94.5 \\ 
\rowcolor{LightGray} BASE: trm-aug-single & 2.0 & 94.5 \\ 
\rowcolor{LightGray} BASE: trm-shared & 2.0 & 86.2 \\ 
cmu\_tartan\_02-1 & 4.3 & 94.0 \\ 
BASE: mono-aug-single & 4.3 & 93.8 \\ 
BASE: mono-shared & 4.3 & 92.0 \\ 
NYU-CUBoulder-03-0 & 4.3 & 91.8 \\ 
cmu\_tartan\_00-1 & 4.3 & 91.8 \\ 
ETHZ-02-1 & 4.3 & 91.8 \\ 
NYU-CUBoulder-04-0 & 4.3 & 83.7 \\ 
\rowcolor{LightGray} uiuc-01-0 & 9.3 & 82.5 \\ 
\rowcolor{LightGray} BASE: trm-single & 9.3 & 82.2 \\ 
IMS-00-0 & 9.3 & 94.3 \\ 
\rowcolor{LightGray} ETHZ-00-1 & 10.3 & 81.7 \\ 
cmu\_tartan\_01-0 & 10.0 & 94.0 \\ 
cmu\_tartan\_01-1 & 10.0 & 93.8 \\ 
\rowcolor{LightGray} cmu\_tartan\_00-0 & 13.0 & 91.9 \\ 
NYU-CUBoulder-02-0 & 10.0 & 91.8 \\ 
\rowcolor{LightGray} flexica-03-1 & 11.7 & 87.2 \\ 
BASE: mono-single & 14.0 & 62.7 \\ 
\rowcolor{LightGray} \textit{*CU7565-01-0} & \textit{20.3} & \textit{93.8} \\ 
BASE: mono-aug-shared & 14.7 & 93.3 \\ 
NYU-CUBoulder-01-0 & 14.7 & 91.4 \\ 
\rowcolor{LightGray} CU7565-02-0 & 17.7 & 90.9 \\ 
LTI-00-1 & 18.3 & 86.2 \\ 
\rowcolor{LightGray} flexica-01-1 & 19.3 & 77.5 \\ 
flexica-02-1 & 20.0 & 70.6 \\ 
\midrule
Oracle (Baselines) & & 97.3 \\
Oracle (Submissions) & & 97.5 \\
Oracle (All) & & 97.7 \\
\bottomrule
\end{tabular}
%\caption{Results on the Iranian genus (3 languages).}
%\label{tab:Iranian}
%\end{table}

 \caption{Results on the Iranian genus (3 languages)}
 \end{subtable}%
 \begin{subtable}{.5\linewidth}
  \centering
%\begin{table}[t]
%\centering
\begin{tabular}{c|c|c}
\toprule
System & Rank & Acc \\ \midrule
deepspin-02-1 & 1.0 & 99.3 \\ 
BASE: trm-single & 1.0 & 99.2 \\ 
deepspin-01-1 & 1.0 & 99.1 \\ 
uiuc-01-0 & 1.0 & 98.7 \\ 
\rowcolor{LightGray} BASE: trm-aug-single & 2.5 & 98.7 \\ 
CULing-01-0 & 3.8 & 99.1 \\ 
\rowcolor{LightGray} cmu\_tartan\_00-0 & 4.4 & 98.0 \\ 
BASE: mono-shared & 7.1 & 97.0 \\ 
\rowcolor{LightGray} NYU-CUBoulder-04-0 & 4.9 & 98.8 \\ 
cmu\_tartan\_01-0 & 6.4 & 98.2 \\ 
\rowcolor{LightGray} BASE: trm-shared & 8.0 & 96.6 \\ 
BASE: mono-aug-shared & 10.4 & 97.6 \\ 
\rowcolor{LightGray} cmu\_tartan\_00-1 & 7.4 & 97.9 \\ 
cmu\_tartan\_01-1 & 9.0 & 98.1 \\ 
\rowcolor{LightGray} NYU-CUBoulder-03-0 & 9.8 & 98.9 \\ 
\rowcolor{LightGray} NYU-CUBoulder-01-0 & 9.8 & 98.6 \\ 
NYU-CUBoulder-02-0 & 10.2 & 98.5 \\ 
\rowcolor{LightGray} BASE: mono-aug-single & 10.2 & 97.5 \\ 
BASE: mono-single & 11.5 & 95.5 \\ 
\rowcolor{LightGray} cmu\_tartan\_02-1 & 10.5 & 97.8 \\ 
BASE: trm-aug-shared & 14.5 & 97.2 \\ 
\rowcolor{LightGray} flexica-03-1 & 17.2 & 93.1 \\ 
IMS-00-0 & 19.0 & 97.6 \\ 
\rowcolor{LightGray} LTI-00-1 & 20.4 & 96.3 \\ 
CU7565-02-0 & 23.1 & 92.9 \\ 
\rowcolor{LightGray} flexica-02-1 & 21.2 & 92.0 \\ 
flexica-01-1 & 26.9 & 87.1 \\ 
\rowcolor{LightGray} ETHZ-02-1 & 25.1 & 86.1 \\ 
ETHZ-00-1 & 27.5 & 81.4 \\ 
\rowcolor{LightGray} \textit{*CU7565-01-0} & \textit{30.0} & \textit{0.0} \\ 
\midrule
Oracle (Baselines) & & 99.4 \\
Oracle (Submissions) & & 99.7 \\
Oracle (All) & & 99.7 \\
\bottomrule
\end{tabular}
%\caption{Results on the Romance genus (8 languages).}
%\label{tab:Romance}
%\end{table}

  \caption{Results on the Romance genus (8 languages)}
 \end{subtable}
\end{table}

\begin{table}[!htb]
\caption{\textbf{Results per Language Genus (in Niger-Congo family)}}
\centering
 \begin{subtable}{.5\linewidth}
  \centering
%\begin{table}[t]
%\centering
\begin{tabular}{c|c|c}
\toprule
System & Rank & Acc \\ \midrule
uiuc-01-0 & 1.0 & 97.7 \\ 
IMS-00-0 & 1.0 & 97.6 \\ 
CULing-01-0 & 1.0 & 96.9 \\ 
\rowcolor{LightGray} NYU-CUBoulder-01-0 & 1.4 & 97.9 \\ 
\rowcolor{LightGray} NYU-CUBoulder-02-0 & 1.4 & 97.9 \\ 
\rowcolor{LightGray} NYU-CUBoulder-03-0 & 1.4 & 97.9 \\ 
\rowcolor{LightGray} deepspin-02-1 & 1.4 & 97.6 \\ 
\rowcolor{LightGray} BASE: mono-aug-single & 1.4 & 97.5 \\ 
\rowcolor{LightGray} BASE: trm-single & 1.4 & 97.4 \\ 
\rowcolor{LightGray} deepspin-01-1 & 1.4 & 97.3 \\ 
\rowcolor{LightGray} NYU-CUBoulder-04-0 & 1.4 & 97.3 \\ 
\rowcolor{LightGray} LTI-00-1 & 1.4 & 97.3 \\ 
\rowcolor{LightGray} BASE: trm-shared & 1.4 & 97.2 \\ 
\rowcolor{LightGray} BASE: mono-single & 1.4 & 97.1 \\ 
\rowcolor{LightGray} BASE: trm-aug-single & 1.4 & 97.0 \\ 
\rowcolor{LightGray} BASE: mono-shared & 1.4 & 97.0 \\ 
\rowcolor{LightGray} BASE: trm-aug-shared & 1.4 & 96.7 \\ 
\rowcolor{LightGray} BASE: mono-aug-shared & 1.4 & 96.6 \\ 
\textit{*CU7565-01-0} & \textit{4.6} & \textit{97.4} \\ 
\rowcolor{LightGray} flexica-02-1 & 3.6 & 96.2 \\ 
\rowcolor{LightGray} flexica-03-1 & 3.6 & 96.2 \\ 
CU7565-02-0 & 4.2 & 95.8 \\ 
\rowcolor{LightGray} cmu\_tartan\_01-1 & 4.2 & 95.6 \\ 
\rowcolor{LightGray} cmu\_tartan\_01-0 & 4.2 & 95.5 \\ 
\rowcolor{LightGray} cmu\_tartan\_00-0 & 4.2 & 95.5 \\ 
cmu\_tartan\_00-1 & 6.5 & 94.9 \\ 
\rowcolor{LightGray} flexica-01-1 & 7.9 & 93.4 \\ 
cmu\_tartan\_02-1 & 13.8 & 93.3 \\ 
\rowcolor{LightGray} ETHZ-02-1 & 16.9 & 92.0 \\ 
ETHZ-00-1 & 18.2 & 89.6 \\ 
\midrule
Oracle (Baselines) & & 98.9 \\
Oracle (Submissions) & & 99.3 \\
Oracle (All) & & 99.5 \\
\bottomrule
\end{tabular}
%\caption{Results on the Bantoid genus (8 languages).}
%\label{tab:Bantoid}
%\end{table}

 \caption{Results on the Bantoid genus (8 languages)}
 \end{subtable}%
 \begin{subtable}{.5\linewidth}
  \centering
%\begin{table}[t]
%\centering
\begin{tabular}{c|c|c}
\toprule
System & Rank & Acc \\ \midrule
BASE: mono-shared & 1.0 & 100.0 \\ 
BASE: mono-single & 1.0 & 100.0 \\ 
CU7565-01-0 & 1.0 & 100.0 \\ 
IMS-00-0 & 1.0 & 100.0 \\ 
deepspin-02-1 & 1.0 & 100.0 \\ 
deepspin-01-1 & 1.0 & 100.0 \\ 
flexica-03-1 & 1.0 & 99.9 \\ 
BASE: trm-shared & 1.0 & 99.9 \\ 
BASE: mono-aug-single & 1.0 & 99.9 \\ 
cmu\_tartan\_00-0 & 1.0 & 99.9 \\ 
BASE: trm-aug-shared & 1.0 & 99.9 \\ 
BASE: trm-aug-single & 1.0 & 99.7 \\ 
cmu\_tartan\_01-1 & 1.0 & 99.7 \\ 
BASE: mono-aug-shared & 1.0 & 99.6 \\ 
NYU-CUBoulder-04-0 & 1.0 & 99.6 \\ 
LTI-00-1 & 1.0 & 99.5 \\ 
flexica-02-1 & 1.0 & 99.3 \\ 
cmu\_tartan\_01-0 & 1.0 & 99.3 \\ 
BASE: trm-single & 1.0 & 98.8 \\ 
NYU-CUBoulder-01-0 & 1.0 & 98.8 \\ 
NYU-CUBoulder-02-0 & 1.0 & 98.8 \\ 
NYU-CUBoulder-03-0 & 1.0 & 98.7 \\ 
cmu\_tartan\_02-1 & 1.0 & 98.7 \\ 
uiuc-01-0 & 1.0 & 98.5 \\ 
\rowcolor{LightGray} CULing-01-0 & 13.0 & 98.0 \\ 
\rowcolor{LightGray} cmu\_tartan\_00-1 & 13.0 & 97.7 \\ 
flexica-01-1 & 14.5 & 97.4 \\ 
\rowcolor{LightGray} CU7565-02-0 & 15.5 & 94.9 \\ 
ETHZ-02-1 & 27.0 & 90.4 \\ 
\rowcolor{LightGray} ETHZ-00-1 & 28.5 & 87.9 \\ 
\midrule
Oracle (Baselines) & & 100.0 \\
Oracle (Submissions) & & 100.0 \\
Oracle (All) & & 100.0 \\
\bottomrule
\end{tabular}
%\caption{Results on the Kwa genus (2 languages).}
%\label{tab:Kwa}
%\end{table}

  \caption{Results on the Kwa genus (2 languages)}
 \end{subtable}
\end{table}

\begin{table}[!htb]
\caption{\textbf{Results per Language Genus (in Oto-Manguean Family)}}
\centering
 \begin{subtable}{.5\linewidth}
  \centering
%\begin{table}[t]
%\centering
\begin{tabular}{c|c|c}
\toprule
System & Rank & Acc \\ \midrule
CULing-01-0 & 1.0 & 93.9 \\ 
uiuc-01-0 & 1.0 & 93.5 \\ 
BASE: trm-single & 1.0 & 92.8 \\ 
\rowcolor{LightGray} deepspin-01-1 & 2.5 & 93.1 \\ 
\rowcolor{LightGray} NYU-CUBoulder-04-0 & 2.5 & 93.1 \\ 
\rowcolor{LightGray} deepspin-02-1 & 2.5 & 92.6 \\ 
\rowcolor{LightGray} NYU-CUBoulder-03-0 & 2.5 & 92.5 \\ 
NYU-CUBoulder-02-0 & 6.0 & 92.3 \\ 
BASE: mono-single & 6.0 & 92.1 \\ 
NYU-CUBoulder-01-0 & 6.0 & 92.0 \\ 
BASE: mono-aug-single & 6.0 & 91.6 \\ 
BASE: trm-aug-single & 6.0 & 91.4 \\ 
\rowcolor{LightGray} IMS-00-0 & 10.5 & 91.4 \\ 
\rowcolor{LightGray} BASE: mono-aug-shared & 10.5 & 90.0 \\ 
\rowcolor{LightGray} BASE: mono-shared & 10.5 & 89.9 \\ 
LTI-00-1 & 13.0 & 89.6 \\ 
cmu\_tartan\_00-1 & 13.0 & 87.9 \\ 
\rowcolor{LightGray} ETHZ-02-1 & 15.5 & 89.7 \\ 
\rowcolor{LightGray} BASE: trm-shared & 15.5 & 89.5 \\ 
cmu\_tartan\_02-1 & 18.0 & 87.3 \\ 
cmu\_tartan\_00-0 & 18.0 & 87.1 \\ 
\rowcolor{LightGray} cmu\_tartan\_01-1 & 20.5 & 86.7 \\ 
cmu\_tartan\_01-0 & 18.0 & 86.3 \\ 
\rowcolor{LightGray} BASE: trm-aug-shared & 21.0 & 84.2 \\ 
ETHZ-00-1 & 22.0 & 82.7 \\ 
\rowcolor{LightGray} \textit{*CU7565-01-0} & \textit{28.0} & \textit{81.7} \\ 
CU7565-02-0 & 26.5 & 76.3 \\ 
\rowcolor{LightGray} flexica-02-1 & 26.5 & 69.2 \\ 
flexica-03-1 & 28.0 & 66.1 \\ 
\rowcolor{LightGray} flexica-01-1 & 29.5 & 40.9 \\ 
\midrule
Oracle (Baselines) & & 96.4 \\
Oracle (Submissions) & & 97.1 \\
Oracle (All) & & 97.4 \\
\bottomrule
\end{tabular}
%\caption{Results on the Amuzgo-Mixtecan genus (2 languages).}
%\label{tab:AmuzgoMixtecan}
%\end{table}

 \caption{Results on the Amuzgo-Mixtecan genus (2 languages)}
 \end{subtable}%
 \begin{subtable}{.5\linewidth}
  \centering
%\begin{table}[t]
%\centering
\begin{tabular}{c|c|c}
\toprule
System & Rank & Acc \\ \midrule
uiuc-01-0 & 1.0 & 81.1 \\ 
\rowcolor{LightGray} CULing-01-0 & 1.5 & 80.3 \\ 
BASE: trm-single & 3.5 & 78.9 \\ 
\rowcolor{LightGray} deepspin-02-1 & 2.2 & 78.7 \\ 
\rowcolor{LightGray} deepspin-01-1 & 2.2 & 78.3 \\ 
\rowcolor{LightGray} NYU-CUBoulder-04-0 & 2.2 & 77.2 \\ 
IMS-00-0 & 3.8 & 78.0 \\ 
NYU-CUBoulder-02-0 & 3.8 & 77.1 \\ 
NYU-CUBoulder-03-0 & 3.8 & 77.0 \\ 
\rowcolor{LightGray} LTI-00-1 & 6.8 & 73.9 \\ 
NYU-CUBoulder-01-0 & 4.8 & 77.5 \\ 
\rowcolor{LightGray} BASE: mono-aug-single & 8.2 & 73.8 \\ 
BASE: mono-aug-shared & 9.2 & 72.9 \\ 
\rowcolor{LightGray} cmu\_tartan\_01-1 & 12.0 & 69.2 \\ 
cmu\_tartan\_00-0 & 13.0 & 68.5 \\ 
cmu\_tartan\_02-1 & 13.0 & 68.5 \\ 
\rowcolor{LightGray} BASE: trm-aug-shared & 15.2 & 65.9 \\ 
BASE: mono-shared & 11.2 & 73.5 \\ 
\rowcolor{LightGray} flexica-01-1 & 21.8 & 51.0 \\ 
BASE: trm-aug-single & 9.2 & 75.7 \\ 
\rowcolor{LightGray} ETHZ-02-1 & 15.0 & 71.7 \\ 
CU7565-02-0 & 16.5 & 68.5 \\ 
\rowcolor{LightGray} BASE: trm-shared & 15.2 & 71.0 \\ 
\rowcolor{LightGray} BASE: mono-single & 15.2 & 70.4 \\ 
cmu\_tartan\_00-1 & 16.5 & 68.9 \\ 
\rowcolor{LightGray} cmu\_tartan\_01-0 & 17.5 & 66.5 \\ 
\textit{*CU7565-01-0} & \textit{26.2} & \textit{75.7} \\ 
\rowcolor{LightGray} ETHZ-00-1 & 26.2 & 60.5 \\ 
flexica-02-1 & 27.0 & 54.3 \\ 
\rowcolor{LightGray} flexica-03-1 & 28.2 & 49.0 \\ 
\midrule
Oracle (Baselines) & & 89.9 \\
Oracle (Submissions) & & 93.7 \\
Oracle (All) & & 94.3 \\
\bottomrule
\end{tabular}
%\caption{Results on the Zapotecan genus (4 languages).}
%\label{tab:Zapotecan}
%\end{table}

  \caption{Results on the Zapotecan genus (4 languages)}
 \end{subtable}
\end{table}

\begin{table}[!htb]
\caption{\textbf{Results per Language Genus (in Oto-Manguean and Uralic Families)}}
\centering
 \begin{subtable}{.5\linewidth}
  \centering
%\begin{table}[t]
%\centering
\begin{tabular}{c|c|c}
\toprule
System & Rank & Acc \\ \midrule
BASE: mono-shared & 1.0 & 98.6 \\ 
uiuc-01-0 & 1.0 & 98.6 \\ 
deepspin-02-1 & 1.0 & 98.5 \\ 
BASE: trm-single & 1.0 & 98.4 \\ 
BASE: mono-single & 1.0 & 98.4 \\ 
BASE: mono-aug-single & 1.0 & 98.4 \\ 
deepspin-01-1 & 1.0 & 98.4 \\ 
\rowcolor{LightGray} BASE: mono-aug-shared & 8.0 & 98.2 \\ 
\rowcolor{LightGray} BASE: trm-aug-single & 8.0 & 98.1 \\ 
CULing-01-0 & 9.5 & 97.7 \\ 
\rowcolor{LightGray} LTI-00-1 & 11.5 & 97.2 \\ 
cmu\_tartan\_01-1 & 12.0 & 96.2 \\ 
\rowcolor{LightGray} cmu\_tartan\_00-1 & 12.0 & 96.8 \\ 
\rowcolor{LightGray} cmu\_tartan\_00-0 & 12.0 & 96.7 \\ 
NYU-CUBoulder-04-0 & 13.5 & 96.5 \\ 
\rowcolor{LightGray} cmu\_tartan\_02-1 & 14.0 & 96.3 \\ 
ETHZ-02-1 & 15.5 & 95.9 \\ 
\rowcolor{LightGray} BASE: trm-shared & 16.5 & 94.2 \\ 
NYU-CUBoulder-03-0 & 18.5 & 94.1 \\ 
NYU-CUBoulder-02-0 & 18.5 & 94.1 \\ 
\rowcolor{LightGray} NYU-CUBoulder-01-0 & 20.0 & 93.7 \\ 
flexica-03-1 & 21.0 & 93.1 \\ 
\rowcolor{LightGray} flexica-02-1 & 22.5 & 93.1 \\ 
cmu\_tartan\_01-0 & 20.5 & 91.9 \\ 
\rowcolor{LightGray} CU7565-02-0 & 25.0 & 91.1 \\ 
IMS-00-0 & 24.5 & 91.0 \\ 
\rowcolor{LightGray} \textit{*CU7565-01-0} & \textit{28.5} & \textit{90.9} \\ 
BASE: trm-aug-shared & 25.5 & 87.3 \\ 
\rowcolor{LightGray} ETHZ-00-1 & 27.5 & 85.3 \\ 
flexica-01-1 & 29.5 & 64.2 \\ 
\midrule
Oracle (Baselines) & & 99.7 \\
Oracle (Submissions) & & 99.9 \\
Oracle (All) & & 99.9 \\
\bottomrule
\end{tabular}
%\caption{Results on the Otomian genus (2 languages).}
%\label{tab:Otomian}
%\end{table}

 \caption{Results on the Otomian genus (2 languages)}
 \end{subtable}%
 \begin{subtable}{.5\linewidth}
  \centering
%\begin{table}[t]
%\centering
\begin{tabular}{c|c|c}
\toprule
System & Rank & Acc \\ \midrule
deepspin-02-1 & 2.2 & 87.4 \\ 
\rowcolor{LightGray} uiuc-01-0 & 2.6 & 83.5 \\ 
deepspin-01-1 & 3.8 & 85.8 \\ 
\rowcolor{LightGray} BASE: trm-aug-single & 4.0 & 84.1 \\ 
BASE: trm-single & 4.3 & 83.4 \\ 
\rowcolor{LightGray} CULing-01-0 & 5.2 & 84.6 \\ 
NYU-CUBoulder-04-0 & 7.0 & 83.0 \\ 
\rowcolor{LightGray} NYU-CUBoulder-02-0 & 10.0 & 82.8 \\ 
NYU-CUBoulder-03-0 & 9.8 & 82.2 \\ 
\rowcolor{LightGray} IMS-00-0 & 12.3 & 82.2 \\ 
NYU-CUBoulder-01-0 & 12.0 & 82.4 \\ 
\rowcolor{LightGray} cmu\_tartan\_00-1 & 8.3 & 80.0 \\ 
cmu\_tartan\_02-1 & 8.3 & 80.2 \\ 
\rowcolor{LightGray} LTI-00-1 & 12.3 & 81.9 \\ 
cmu\_tartan\_01-1 & 8.0 & 80.3 \\ 
\rowcolor{LightGray} cmu\_tartan\_00-0 & 9.4 & 80.8 \\ 
BASE: trm-aug-shared & 18.9 & 76.9 \\ 
\rowcolor{LightGray} CU7565-02-0 & 20.3 & 74.0 \\ 
\textit{*CU7565-01-0} & \textit{27.1} & \textit{92.9} \\ 
\rowcolor{LightGray} BASE: mono-single & 12.6 & 75.5 \\ 
cmu\_tartan\_01-0 & 11.7 & 78.6 \\ 
\rowcolor{LightGray} BASE: mono-shared & 15.8 & 74.8 \\ 
BASE: mono-aug-shared & 16.9 & 77.4 \\ 
\rowcolor{LightGray} BASE: trm-shared & 21.2 & 67.3 \\ 
ETHZ-02-1 & 20.6 & 61.0 \\ 
\rowcolor{LightGray} BASE: mono-aug-single & 11.2 & 80.7 \\ 
flexica-02-1 & 21.2 & 57.3 \\ 
\rowcolor{LightGray} flexica-03-1 & 23.0 & 52.5 \\ 
flexica-01-1 & 26.6 & 56.1 \\ 
\rowcolor{LightGray} ETHZ-00-1 & 28.2 & 45.7 \\ 
\midrule
Oracle (Baselines) & & 93.9 \\
Oracle (Submissions) & & 95.8 \\
Oracle (All) & & 96.3 \\
\bottomrule
\end{tabular}
%\caption{Results on the Finnic genus (10 languages).}
%\label{tab:Finnic}
%\end{table}

  \caption{Results on the Finnic genus (10 languages)}
 \end{subtable}
\end{table}

\begin{table}[!htb]
\caption{\textbf{Results per Language Genus (in Uralic Family)}}
\centering
 \begin{subtable}{.5\linewidth}
  \centering
%\begin{table}[t]
%\centering
\begin{tabular}{c|c|c}
\toprule
System & Rank & Acc \\ \midrule
deepspin-01-1 & 1.0 & 97.9 \\ 
deepspin-02-1 & 1.0 & 97.9 \\ 
\rowcolor{LightGray} CULing-01-0 & 2.0 & 97.8 \\ 
BASE: trm-single & 3.0 & 97.7 \\ 
\rowcolor{LightGray} cmu\_tartan\_00-1 & 5.0 & 97.4 \\ 
uiuc-01-0 & 5.0 & 97.6 \\ 
BASE: trm-aug-single & 5.0 & 97.6 \\ 
\rowcolor{LightGray} cmu\_tartan\_00-0 & 6.0 & 97.4 \\ 
\rowcolor{LightGray} cmu\_tartan\_01-1 & 6.0 & 97.3 \\ 
cmu\_tartan\_02-1 & 12.5 & 95.6 \\ 
\rowcolor{LightGray} cmu\_tartan\_01-0 & 9.0 & 97.1 \\ 
BASE: mono-single & 9.5 & 97.0 \\ 
\rowcolor{LightGray} BASE: mono-aug-single & 11.0 & 96.7 \\ 
NYU-CUBoulder-04-0 & 14.0 & 95.6 \\ 
\rowcolor{LightGray} LTI-00-1 & 13.5 & 96.7 \\ 
BASE: trm-shared & 14.5 & 95.7 \\ 
\rowcolor{LightGray} BASE: trm-aug-shared & 17.0 & 95.6 \\ 
flexica-02-1 & 18.5 & 95.0 \\ 
\rowcolor{LightGray} NYU-CUBoulder-02-0 & 18.5 & 94.8 \\ 
IMS-00-0 & 19.0 & 94.8 \\ 
\rowcolor{LightGray} NYU-CUBoulder-03-0 & 18.5 & 94.8 \\ 
\rowcolor{LightGray} NYU-CUBoulder-01-0 & 18.5 & 94.7 \\ 
flexica-03-1 & 19.0 & 94.6 \\ 
\rowcolor{LightGray} BASE: mono-shared & 21.0 & 94.5 \\ 
CU7565-02-0 & 23.5 & 93.3 \\ 
\rowcolor{LightGray} BASE: mono-aug-shared & 26.0 & 91.5 \\ 
flexica-01-1 & 27.0 & 88.7 \\ 
\rowcolor{LightGray} ETHZ-02-1 & 28.0 & 79.4 \\ 
ETHZ-00-1 & 29.0 & 73.4 \\ 
\rowcolor{LightGray} \textit{*CU7565-01-0} & \textit{30.0} & \textit{0.0} \\ 
\midrule
Oracle (Baselines) & & 98.6 \\
Oracle (Submissions) & & 99.0 \\
Oracle (All) & & 99.2 \\
\bottomrule
\end{tabular}
%\caption{Results on the Permic genus (2 languages).}
%\label{tab:Permic}
%\end{table}

 \caption{Results on the Permic genus (2 languages)}
 \end{subtable}%
 \begin{subtable}{.5\linewidth}
  \centering
%\begin{table}[t]
%\centering
\begin{tabular}{c|c|c}
\toprule
System & Rank & Acc \\ \midrule
deepspin-02-1 & 1.0 & 94.0 \\ 
CULing-01-0 & 1.0 & 93.9 \\ 
BASE: trm-single & 1.0 & 93.9 \\ 
uiuc-01-0 & 1.0 & 93.8 \\ 
\rowcolor{LightGray} BASE: trm-aug-single & 3.5 & 93.7 \\ 
\rowcolor{LightGray} deepspin-01-1 & 3.5 & 93.6 \\ 
cmu\_tartan\_02-1 & 6.5 & 93.3 \\ 
cmu\_tartan\_00-1 & 6.5 & 93.2 \\ 
cmu\_tartan\_01-1 & 6.5 & 93.2 \\ 
cmu\_tartan\_01-0 & 6.5 & 93.2 \\ 
cmu\_tartan\_00-0 & 6.5 & 93.2 \\ 
\rowcolor{LightGray} BASE: mono-single & 9.5 & 93.0 \\ 
\rowcolor{LightGray} LTI-00-1 & 9.5 & 92.8 \\ 
BASE: trm-shared & 13.5 & 92.0 \\ 
\rowcolor{LightGray} BASE: mono-aug-single & 14.5 & 92.3 \\ 
BASE: trm-aug-shared & 15.0 & 91.9 \\ 
\rowcolor{LightGray} IMS-00-0 & 17.0 & 91.5 \\ 
NYU-CUBoulder-04-0 & 18.5 & 90.8 \\ 
\rowcolor{LightGray} flexica-03-1 & 18.5 & 90.5 \\ 
\rowcolor{LightGray} flexica-02-1 & 18.5 & 90.5 \\ 
NYU-CUBoulder-03-0 & 19.5 & 90.2 \\ 
NYU-CUBoulder-02-0 & 19.5 & 90.2 \\ 
\rowcolor{LightGray} NYU-CUBoulder-01-0 & 23.5 & 89.5 \\ 
BASE: mono-shared & 21.5 & 88.9 \\ 
\rowcolor{LightGray} BASE: mono-aug-shared & 24.5 & 87.2 \\ 
CU7565-02-0 & 25.5 & 85.2 \\ 
\rowcolor{LightGray} flexica-01-1 & 27.0 & 82.1 \\ 
ETHZ-02-1 & 28.0 & 73.7 \\ 
\rowcolor{LightGray} ETHZ-00-1 & 28.5 & 67.9 \\ 
\textit{*CU7565-01-0} & \textit{30.0} & \textit{0.0} \\ 
\midrule
Oracle (Baselines) & & 97.0 \\
Oracle (Submissions) & & 97.6 \\
Oracle (All) & & 98.0 \\
\bottomrule
\end{tabular}
%\caption{Results on the Mordvin genus (2 languages).}
%\label{tab:Mordvin}
%\end{table}

  \caption{Results on the Mordvin genus (2 languages)}
 \end{subtable}
\end{table}

%\pagebreak
%\section{Results per Genus}
%\edo{Table captions should read `genus' rather than `family'.}

%\subsection{Afro-Asiatic}
%\input{tables/Semitic}
%\pagebreak

%\subsection{Indo-European}

%\input{tables/Iranian}
%\input{tables/Romance}

%\pagebreak
%\subsection{Niger-Congo}
%\input{tables/Bantoid}
%\input{tables/Kwa}

%\pagebreak
%\subsection{Oto-Manguean}
%\input{tables/AmuzgoMixtecan}
%\input{tables/Zapotecan}
%\input{tables/Otomian}

%\pagebreak
%\subsection{Uralic}
%\input{tables/Finnic}
%\input{tables/Permic}
%\input{tables/Mordvin}

%\input{tables/}
%\input{tables/}
%\input{tables/}

\end{document}